# On the Robustness of Semantic Segmentation Models to Adversarial Attacks

Anurag Arnab    Ondrej Miksik    Philip H.S. Torr
University of Oxford
{anurag.arnab, ondrej.miksik, philip.torr}@eng.ox.ac.uk

## Abstract

*Deep Neural Networks (DNNs) have demonstrated exceptional performance on most recognition tasks such as image classification and segmentation. However, they have also been shown to be vulnerable to adversarial examples. This phenomenon has recently attracted a lot of attention but it has not been extensively studied on multiple, large-scale datasets and structured prediction tasks such as semantic segmentation which often require more specialised networks with additional components such as CRFs, dilated convolutions, skip-connections and multiscale processing.*

*In this paper, we present what to our knowledge is the first rigorous evaluation of adversarial attacks on modern semantic segmentation models, using two large-scale datasets. We analyse the effect of different network architectures, model capacity and multiscale processing, and show that many observations made on the task of classification do not always transfer to this more complex task. Furthermore, we show how mean-field inference in deep structured models, multiscale processing (and more generally, input transformations) naturally implement recently proposed adversarial defenses. Our observations will aid future efforts in understanding and defending against adversarial examples. Moreover, in the shorter term, we show how to effectively benchmark robustness and show which segmentation models should currently be preferred in safety-critical applications due to their inherent robustness.*

## 1. Introduction

Computer vision has progressed to the point where Deep Neural Network (DNN) models for most recognition tasks such as classification or segmentation have become a widely available commodity. State-of-the-art performance on various datasets has increased at an unprecedented pace, and as a result, these models are now being deployed in more and more complex systems. However, despite DNNs performing exceptionally well in absolute performance scores, they have also been shown to be vulnerable to *adversarial examples* – images which are classified incorrectly (often with high confidence), although there is only a minimal perceptual difference with correctly classified inputs [28, 10, 83].

This raises doubts about DNNs being used in safety-critical applications such as driverless vehicles [47] or medical diagnosis [30] since the networks could inexplicably classify a natural input incorrectly although it is almost identical to examples it has classified correctly before (Fig. 1). Moreover, it allows for the possibility of malicious agents attacking systems that use neural networks [54, 70, 77, 32]. Hence, the robustness of networks perturbed by adversarial noise may be as important as the predictive accuracy on clean inputs. And if multiple models achieve comparable performance, we should always consider deploying the one which is inherently most robust to adversarial examples in (safety-critical) production settings.

This phenomenon has recently attracted a lot of attention and numerous strategies have been proposed to train DNNs to be more robust to adversarial examples [38, 55, 73, 63]. However, these defenses are not universal; they have frequently been found to be vulnerable to other types of attacks [18, 16, 17, 44] and/or come at the cost of performance penalties on clean inputs [19, 40, 63]. To the best of our knowledge, adversarial examples have not been extensively analysed beyond standard image classification models, and often on small datasets such as MNIST or CIFAR-10 [63, 40, 73]. Hence, the vulnerability of modern DNNs to adversarial attacks on more complex tasks such as semantic segmentation in the context of real-world datasets covering different domains remains unclear.

In this paper, we present what to our knowledge is the first rigorous evaluation of the robustness of semantic segmentation models to adversarial attacks. We focus on semantic segmentation, since it is a significantly more complex task than image classification [8]. This has also been witnessed by the fact that state-of-the-art semantic segmentation models are typically based on standard image classification architectures [53, 80, 43], extended by additional components such as dilated convolutions [21, 90], specialised pooling [22, 92], skip-connections [60], Conditional Random Fields (CRFs) [93, 1] and/or multiscale processing [22, 20] whose impact on the robustness has never

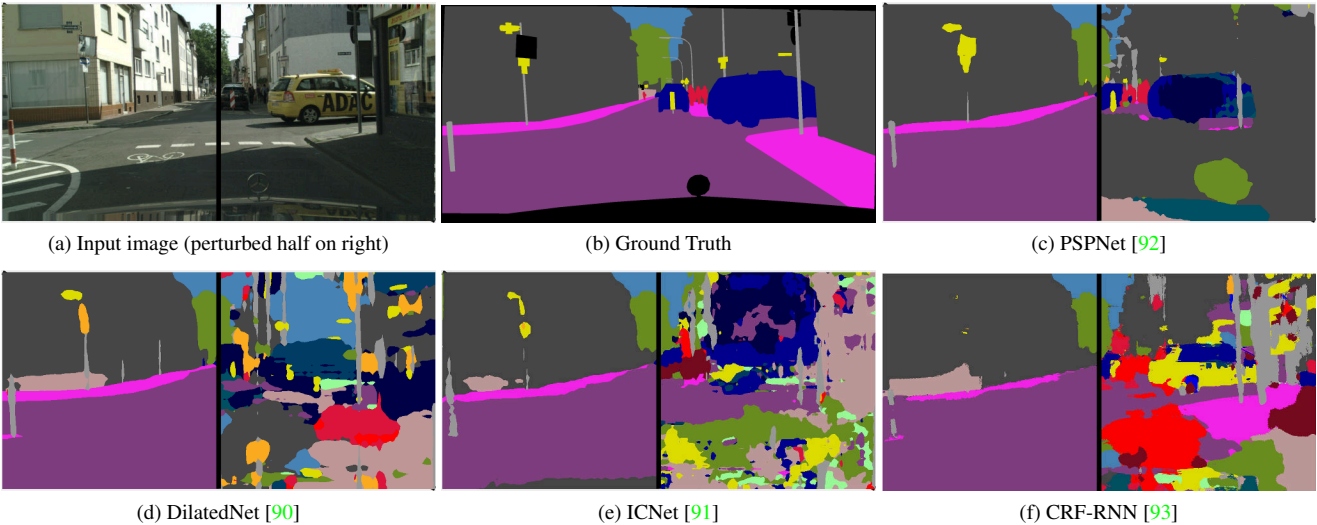

Figure 1: The left hand side shows the original image, and the right the output when modified with imperceptible adversarial perturbations. There is a large variance in how each network's performance is degraded, even though the perturbations are created individually for each network with the same $\ell_\infty$ norm of 4. We rigorously analyse a diverse range of state-of-the-art segmentation networks, observing how architectural properties, such as residual connections, multiscale processing and CRFs, and input transformations, all influence adversarial robustness. These observations will help future efforts to understand and defend against adversarial examples, whilst in the short term they suggest which networks should currently be preferred in safety-critical applications.

been thoroughly studied.

First, we analyse the robustness of various DNN architectures to adversarial examples and show that the Deeplab v2 network [22] is significantly more robust than approaches which achieve better prediction scores on public benchmarks [92]. Thereafter, we show that adversarial examples are less effective when processed at different scales. Furthermore, multiscale networks are more robust to multiple different attacks and white-box attacks on them produce more transferable perturbations. Inspired by the effect of multiscale processing, we examine other input transformations which neural networks are not invariant to and show that they are markedly more robust to transformed adversarial examples. However, we also show that this is true only when the attack generation process does not take knowledge of these input transformations into account; otherwise, the robustness improvements are rather marginal. These observation have important implications on producing effective physical adversarial examples in the real world. On a separate track, we also show that structured prediction models have a similar effect as "gradient-masking" defense strategies [71, 73]. As such, mean field CRF inference increases robustness to untargeted adversarial attacks, but in contrast to the gradient masking defense, it also improves the network's predictive accuracy. Another of our contributions shows that some widely accepted observations about robustness and model size or iterative attacks, which were made in the context of image classification [55, 63] do not transfer to semantic segmentation and different, real-world datasets. Moreover, we also show that proposed adversarial defenses should be evaluated prudently by using knowledge of the defense mechanism in the white-box attack to test it, which was not done in previously [41, 86, 56, 24]. Finally, in contrast to the prior art [55, 59], our experiments are carried out on two large-scale, real-world datasets and (most of) our observations remained consistent across them.

We believe our findings will facilitate future efforts in understanding and defending against adversarial examples without compromising predictive accuracy.

## 2. Adversarial Examples

Adversarial perturbations cause a classifier to change its original prediction, when added to the original input $\mathbf{x}$. This phenomenon was initially studied in the context of malware detection and spam classification [10, 28], and has more recently become popular in the context of computer vision. For a neural network $f$ parametrised by $\theta$ that maps $\mathbf{x} \in \mathbb{R}^m$ to $y$, a target class from $\mathcal{L} = \{1, 2, \ldots, L\}$, Szegedy et al. [83] defined an adversarial perturbation $\mathbf{r}$ as the solution to the optimisation problem

$$\arg\min \quad \|\mathbf{r}\|_2 \quad \text{subject to} \quad f(\mathbf{x} + \mathbf{r}; \theta) = y_t, \quad (1)$$

where $y_t$ is the target label of the adversarial example $\mathbf{x}^{adv} = \mathbf{x} + \mathbf{r}$. For clarity of exposition, we consider only a single label $y$. This naturally generalises to the case of semantic segmentation where networks are trained with an independent cross-entropy loss at each pixel.

Constraining the neural network to output $y$ is difficult to optimise. Hence, [83] added an additional term to the objective based on the loss function used to train the network

$$\arg\min_{\mathbf{r}} \quad c\|\mathbf{r}\|_2 + L(f(\mathbf{x} + \mathbf{r}; \theta), y_t). \tag{2}$$

Here, $L$ is the loss function between the network prediction and desired target, and $c$ is a positive scalar. Szegedy *et al.* [83] solved this using L-BFGS, and [18] and [23] have proposed further advances using surrogate loss functions. However, this method is computationally very expensive as it requires several minutes to produce a single attack. Hence, the following methods are used in practice:

**Fast Gradient Sign Method (FGSM) [38].** FGSM produces adversarial examples by increasing the loss (usually the cross-entropy) of the network on the input $\mathbf{x}$ as

$$\mathbf{x}^{adv} = \mathbf{x} + \epsilon \cdot \text{sign}(\nabla_\mathbf{x} L(f(\mathbf{x}; \theta), y)). \tag{3}$$

This is a single-step, untargeted attack, which approximately minimises the $\ell_\infty$ norm of the perturbation bounded by the parameter $\epsilon$.

**FGSM ll [55].** This single-step attack encourages the network to classify the adversarial example as $y_t$ by assigning

$$\mathbf{x}^{adv} = \mathbf{x} - \epsilon \cdot \text{sign}(\nabla_\mathbf{x} L(f(\mathbf{x}; \theta), y_t)). \tag{4}$$

We follow the convention of choosing the target class as the least likely class predicted by the network [55].

**Iterative FGSM [55, 63].** This attack extends FGSM by applying it in an iterative manner, which increases the chance of fooling the original network. Using the subscript to denote the iteration number, this can be written as

$$\mathbf{x}_0^{adv} = \mathbf{x} \tag{5}$$
$$\mathbf{x}_{t+1}^{adv} = \text{clip}(\mathbf{x}_t^{adv} + \alpha \cdot \text{sign}(\nabla_{\mathbf{x}_t^{adv}} L(f(\mathbf{x}_t^{adv}; \theta), y)), \epsilon)$$

The $\text{clip}(\mathbf{a}, \epsilon)$ function makes sure that each element $a_i$ of $\mathbf{a}$ is in the range $[a_i - \epsilon, a_i + \epsilon]$. This ensures that the max-norm constraint of each component of the perturbation $\mathbf{r}$, being no greater than $\epsilon$ is maintained. It thus corresponds to projected gradient descent [63], with step-size $\alpha$, into an $\ell_\infty$ ball of radius $\epsilon$ around the input $\mathbf{x}$.

**Iterative FGSM ll [55].** This is a stronger version of FGSM ll. This attack sets the target to be the least likely class predicted by the network, $y_{ll}$, in each iteration

$$\mathbf{x}_{t+1}^{adv} = \text{clip}(\mathbf{x}_t^{adv} - \alpha \cdot \text{sign}(\nabla_{\mathbf{x}_t^{adv}} L(f(\mathbf{x}_t^{adv}; \theta), y_{ll})), \epsilon). \tag{6}$$

The aforementioned attacks were all proposed in the context of image classification, but they have been adapted to the problems of semantic segmentation [35, 23], object detection [87] and visual question answering [89]. Similar, gradient-based attacks have also been proposed to minimise the $\ell_2$ norm of the adversarial perturbation, $\mathbf{r}$, [67, 18], and also to attack other classification algorithms such as SVMs [10]. Methods to optimise the non-differentiable $\ell_0$ norm of the perturbation have also been proposed [82, 72, 69].

## 3. Adversarial Defenses and Evaluations

Liu *et al.* [59] have thoroughly evaluated the transferability of adversarial examples generated on one network and tested on another unknown model, *i.e.* only as "blackbox" attacks [83, 71, 65, 66]. Kurakin *et al.* [55], contrastingly, studied the adversarial training defense, which generates adversarial examples online and adds them into the training set [38, 63, 84]. They found that training with adversarial examples generated by single-step methods conferred robustness to other single-step attacks with negligible performance difference to normally trained networks on clean inputs. However, the adversarially trained network was still as vulnerable to iterative attacks as standard models. Madry *et al.* [63], conversely, found robustness to iterative attacks by adversarial training with them. However, this was only on the small MNIST dataset. The defense was not effective on CIFAR-10, underlining the importance of testing on multiple datasets. Tramer *et al.* [84] also found that adversarially trained models were still susceptible to black-box, single-step attacks generated from other networks. Other adversarial defenses based on detecting the perturbation in the input [64, 39, 34, 88, 81] or pre-processing the input [41, 56, 86, 76] have also all been subverted [5, 4, 17, 44, 16, 85]. Recently, progress has been made on formal verification of neural networks [49, 14] which can provably compute whether adversarial examples with a particular norm exist for a network. However, as these methods are limited to certain architectures, norms which are linear, and do not scale to large networks, they cannot be used on the state-of-the-art networks we consider in this work.

Currently, no effective defense to all adversarial attacks exist. This motivates us, for the first time to our knowledge, to study the properties of state-of-the-art segmentation networks and how they affect robustness to various adversarial attacks. Previous evaluations have only considered standard classification networks (Inception in [55], and GoogleNet, VGG and ResNet in [59]). We consider the more complex task of semantic segmentation, and evaluate eight different architectures, some of them with multiple classification backbones, and show that some features of semantic segmentation models (such as CRFs and multi-scale processing) naturally implement recently proposed adversarial defenses. Moreover, our evaluation is carried out on two large-scale datasets instead of only ImageNet as [55, 59]. This allows us to show that not all previously observed empirical results on classification transfer to segmentation.

The conclusions from our evaluations may thus aid future efforts to develop defenses to adversarial attacks that preserve predictive accuracy. Moreover, our results suggests which state-of-the-art models for semantic segmentation should currently be preferred in (safety-critical) settings where both accuracy and robustness are a priority.

## 4. Experimental Set-up

We describe the datasets, DNN models, adversarial attacks and evaluation metrics used for our evaluation in this section. Exhaustive details are included in the supplementary. We have also released our code[1] to aid reproducibility.

**Datasets.** We use the Pascal VOC [31] and Cityscapes [25] validation sets, the two most widely used semantic segmentation benchmarks. Pascal VOC consists of internet-images labelled with 21 different classes. The reduced validation [93, 60] set contains 346 images, and the training set has about 70000 images when combined with additional annotations from [42] and [58]. Cityscapes consists of road-scenes captured from car-mounted cameras and has 19 classes. The validation set has 500 images, and the training set totals about 23000 images. As this dataset provides high-resolution imagery ($2048 \times 1024$ pixels) which require too much memory for some models, we have resized all images to $1024 \times 512$ when evaluating.

**Models.** We use a wide variety of current or previous state-of-the-art models, ranging from lightweight networks suitable for embedded applications to complex models which explicitly enforce structural constraints. Whenever possible, we have used publicly available code or trained models. The models we had to retrain achieve similar performance to the ones trained by the original authors.

We used the public models of CRF-RNN [93], Dilated-Net [90], PSPNet [92] on Cityscapes, ICNet [92] and Seg-Net [7]. We retrained FCN [60] and E-Net [74], as well as Deeplab v2 [22] and PSPNet for VOC as the public models are trained with the validation set. Our selection of networks are based on both VGG [80] and ResNet [43] backbones, whilst E-Net and ICNet employ custom architectures for real-time applications whose parameters measure only 1.5MB and 30.1MB in 32-bit floats, respectively. Furthermore, the models we evaluate use a variety of unique approaches including dilated convolutions [90, 22], skip-connections [60], specialised pooling [92, 22], encoder-decoder architecture [7, 74], multiscale processing [22] and CRFs [93]. In all our experiments, we evaluate the model in the same manner it was trained – CRF post-processing or multiscale ensembling is not performed unless the network incorporated CRFs [93] or multiscale averaging [22] as network layers whilst training.

[1] www.robots.ox.ac.uk/~aarnab/adversarial_robustness.html

**Adversarial attacks.** We use the FGSM, FGSM ll, Iterative FGSM and Iterative FGSM ll attacks described in Sec. 2. Following [55], we set the number of iterations of iterative attacks to $\min(\epsilon + 4, \lceil 1.25\epsilon \rceil)$ and step-size $\alpha = 1$ meaning that the value of each pixel is changed by 1 every iteration. The Iterative FGSM (untargeted) and FGSM ll (targeted) attacks are only reported in the supplementary as we observed similar trends on FGSM and Iterative FGSM ll. We evaluated these attacks when setting the $\ell_\infty$ norm of the perturbations $\epsilon$ to each value from $\{0.25, 0.5, 1, 2, 4, 8, 16, 32\}$. Even small values such as $\epsilon = 0.25$ introduce errors among all the models we evaluated. The maximum value of $\epsilon$ was chosen as 32 since the perturbation is conspicuous at this point. Qualitative examples of these attacks are shown in the supplementary.

**Evaluation metric.** The Intersection over Union (IoU) is the primary metric used in evaluating semantic segmentation [31, 25]. However, as the accuracy of each model varies, we adapt the relative metric used by [55] for image classification and measure adversarial robustness using the *IoU Ratio* – the ratio of the network's IoU on adversarial examples to that on clean images computed over the entire dataset. As the relative ranking between models for the IoU Ratio and absolute IoU is typically the same, we report the latter only in the supplementary.

## 5. The robustness of different architectures

We evaluate the robustness of different architectures and show how our observations regarding model capacity and single-step attacks do not corroborate with some previous findings in the context of image classification [55, 63]. Additionally, our results also support why JPEG compression as a pre-processing step mitigates small perturbations [29].

### 5.1. The robustness of different networks

Fig. 2 shows the robustness of several state-of-the-art models on the VOC dataset. In general, ResNet-based models not only achieve higher accuracy on clean inputs but are also more robust to adversarial inputs. This is particularly the case for the single-step FGSM attack (Fig. 2a). On the more effective Iterative FGSM ll attack, the margin between the most and least robust network is smaller as none of them perform well (Fig. 2b). However, we note that iterative attacks tend not to transfer to other models [55] (Sec. 6.2).

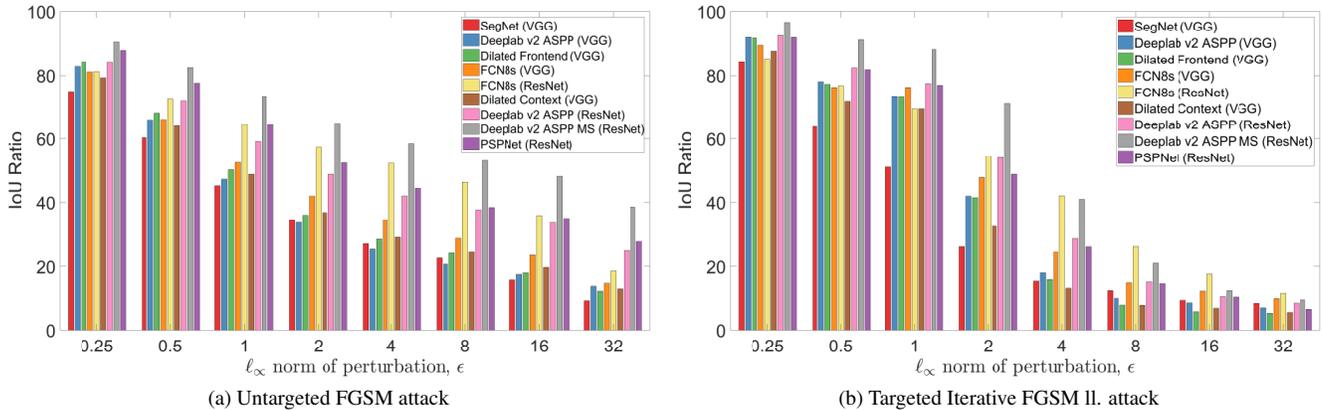

Figure 2: Adversarial robustness of state-of-the-art models on Pascal VOC. Models based on the ResNet backbone tend to be more robust. For instance, FCN8s and Deeplab v2 ASPP with a ResNet-101 backbone are more robust than with the VGG backbone. Moreover, as expected, the Iterative FGSM ll attack is more powerful at fooling networks than single-step FGSM. Models are ordered by increasing IoU on clean inputs. Results on additional attacks are in the supplementary.

Thus, they are less useful in practical, black-box attacks.

In particular, we have evaluated the FCN8s [60] and Deeplab-v2 with ASPP [22] models based on the popular VGG-16 [80] and ResNet-101 [43] backbones. In both cases, the ResNet variant shows greater robustness. We also observe that most of the networks achieve similar scores on clean inputs. As a result, the relative rankings of models in Fig. 2 for the IoU Ratio is about the same as their ranking on clean inputs. Furthermore, the best performing model on clean inputs, PSPNet [92] is actually less robust than Deeplab v2 with Multiscale ASPP [22]: For all $\epsilon$ values we tested, the absolute IoU score of Deeplab v2 was higher than PSPNet. These observations as well as results on FGSM ll and Iterative FGSM showing that the relative ranking of robustness for the different networks is similar, are detailed in the supplementary material.

### 5.2. Model capacity and residual connections

Madry *et al*. [63] and Kurakin *et al*. [55] have studied the effect of model capacity on adversarial robustness by changing the number of filters at each DNN layer, since they used the parameter count as a proxy for model capacity. Madry *et al*. [63] observed on MNIST and CIFAR-10, that networks, trained on clean examples, with a small number of parameters are the most vulnerable to adversarial examples. This observation would have serious safety implications on deployment of lightweight models, typically required by embedded platforms. Instead, we analyse different network structures and show in Fig. 3 that lightweight networks such as E-Net [74] (only 1.5 MB) and IC-Net [91] (only 30.1 MB) are affected by adversarial examples similarly as Dilated-Net [90] which has 512.6 MB in parameters (using 32-bit floats). Dilated-Net is only more robust than both of these lightweight networks for FGSM and FGSM-ll with $\epsilon \geq 4$ (which is also when perturbations become visible to the naked eye). Note that both E-Net and IC-Net have custom backbones and heavily use residual connections.

Fig. 3 also shows that adding the "Context" module of Dilated-Net onto the "Front-end" slightly reduces robustness across all $\epsilon$ values on both attacks on Cityscapes. Fig. 2 shows that this is observed for most $\epsilon$ values on VOC as well. This is even though the additional parameters of the "Context" module increases accuracy on clean inputs. Whilst models with higher capacity may be more resistant to adversarial attacks, one cannot compare the capacities of different networks, given that neither the most accurate network (PSPNet) or the network with the most parameters (Dilated-Net) are actually the most robust.

### 5.3. The unexpected effectiveness of single-step methods on Cityscapes

The single-step FGSM and FGSM ll attacks are significantly more effective on Cityscapes than on Pascal VOC. The IoU ratio for FGSM at $\epsilon = 32$ for PSPNet and Dilated Context is 2.5% and 2.8%, respectively, on Cityscapes. On Pascal VOC, it is substantially higher at 27.9% and 12.2%. Single-step methods (which only search in a 1-D subspace in the space of images) also appear to outperform iterative methods for high $\epsilon$ values on Cityscapes. In contrast, iterative attacks appear about as effective on Cityscapes as on Pascal VOC, when using the same hyperparameters as [55].

Thus, it may be a dataset property that causes the network to learn weights more susceptible to single-step attacks. Cityscapes has, subjectively, less variability than VOC and it also labels "stuff" classes [36]. The effect of the training set on adversarial attacks has not been considered before, and most prior work used MNIST [83, 38, 63] or ImageNet [55, 84, 59]. However, [11] and [51], showed

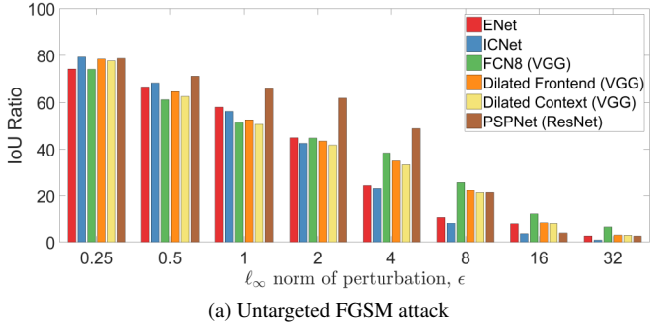
(a) Untargeted FGSM attack

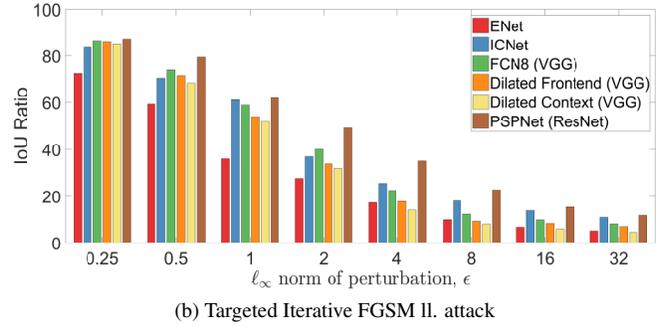
(b) Targeted Iterative FGSM ll. attack

Figure 3: Adversarial robustness of state-of-the-art models on the Cityscapes dataset. Contrary to Madry *et al.* [63], we observe that lightweight networks such as E-Net [74] and ICNet [91] are often about as robust as Dilated-Net [90] (341× more parameters than E-Net). Dilated-Net without its "Context" module is slightly more robust than the full network. As with the VOC dataset, ResNet (PSPNet) architectures are more robust than VGG (Dilated-Net and FCN8). Curiously, the FGSM attack is more effective than Iterative FGSM ll which computes adversarial examples from a larger search space.

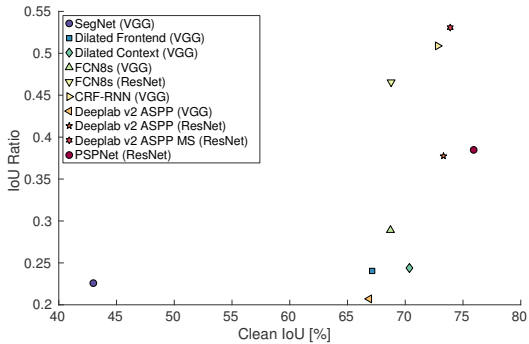

Figure 4: The IoU Ratio compared to the IoU on clean inputs on the Pascal VOC dataset, for the FGSM attack with $\epsilon = 8$. The relative ordering of the models is the same if we plot the absolute IoU on adversarial inputs, with the exception of SegNet which is then ranked the lowest.

that the test error of an SVM and neural network could respectively be increased by inserting "poisonous" examples into its training set. Results from the FGSM ll attack, which shows the same trend as FGSM, are in the supplementary.

### 5.4. Imperceptible perturbations

With $\epsilon = 0.25$, the perturbation is so small that the RGB values of the image pixels (assuming integers $\in [0, 255]$) are usually unchanged. Nevertheless, Fig. 2 and 3 show that the performance of all analysed models were degraded by at least 9% relative IoU for each attack. The observation of [29], that lossy JPEG as a pre-processing step helps to mitigate FGSM for small $\epsilon$ is thus not surprising as JPEG does not entirely preserve these small, high-frequency perturbations and the result is also finally rounded to integers.

### 5.5. Relation with concurrent work

Our results are also corroborated by the concurrent work of Cubuk *et al.* [26] who performed Neural Architecture Search to find architectures that are more robust to adversarial examples. Cubuk *et al.* [26] found that their best architecture had more identity connections and depth than their baseline. This agrees with our observation that models based on ResNet typically have higher robustness and accuracy on clean inputs.

The authors also observed a correlation between accuracy on clean data and robustness. We also observed this correlation (Fig. 4), although the most accurate model on clean inputs (PSPNet) is not the most robust (Deeplab v2 Multiscale). Figure 4 shows the results on the FGSM attack at $\epsilon = 8$, for consistency with [26].

### 5.6. Discussion

We have shown that models with residual connections (ResNet, E-Net, ICNet) are inherently more robust than chain-like VGG-based networks, even if the number of parameters of the VGG model is orders of magnitude larger. Moreover, Dilated-Net, without its "Context" module is more robust than its more performant, full version. This is contrary to the observations regarding parameter count of [63] and [55] who simply increased the number of filters at each layer. The most robust model was Deeplab v2 with Multiscale ASPP, outperforming the current state-of-the-art PSPNet [92], in absolute IoU on adversarial inputs.

We also found that perturbations that do not even change the image's integral RGB values still degraded performance of all models, and that single-step attacks are significantly more effective on Cityscapes than VOC, achieving as low as 0.8% relative IoU. This was unexpected, given that single-step methods only search in a one-dimensional subspace, and raises questions about how the training data of a net-work affects its decision boundaries. Also, explaining the effect of residual connections on adversarial robustness remains an open research question. As Deeplab v2 showed a significant increase in robustness over its single-scale variant, we analyse the effects of multiscale processing next in Sec. 6. Thereafter, we study CRFs, a common component in semantic segmentation models.

## 6. Multiscale Processing and Transferability of Adversarial Examples

Deeplab v2 with Multiscale ASPP was the most robust model to various attacks in Sec. 5, with a significant difference to its single-scale variant. In this section, we first examine the effect of multiscale processing and then relate our observations to concurrent work.

### 6.1. Multiscale processing

The Deeplab v2 network processes images at three different resolutions, 50%, 75% and 100% where the weights are shared among each of the scale branches. The results from each scale are upsampled to a common resolution, and then max-pooled such that the most confident prediction at each pixel from each of the scale branches is chosen [22]. This network is trained in this multiscale manner, although it is possible to perform this multiscale ensembling as a post-processing step at test-time only [21, 27, 57, 92].

We hypothesise that adversarial attacks, when generated at a single scale, are no longer as malignant when processed at another. This is because CNNs are not invariant to scale, and a range of other transformations [33, 75, 45]. And although it is possible to generate adversarial attacks from multiple different scales of the input, these examples may not be as effective at a single scale, making networks which process images at multiple scales more robust. We investigate the transferability of adversarial perturbations generated at one scale and evaluated at another in Sec. 6.2, and the robustness and transferability of multiscale networks in Sec. 6.3. Thereafter, we relate our findings to concurrent work.

### 6.2. The transferability of adversarial examples at different scales

Table 1 shows results for the FGSM and Iterative FGSM ll attacks. The diagonals show "white-box" attacks where the adversarial examples are generated from the attacked network. These attacks typically result in the greatest performance degradation, as expected. The off-diagonals show the transferability of perturbations generated from other networks. In constrast to Iterative FGSM ll, FGSM attacks transfer well to other networks, which confirms the observations [55] made in the context of image classification.

The attack produced from 50% resolution inputs transfers poorly to other scales of Deeplab v2 and other architectures, and vice versa. This is seen by looking across the columns and rows of Tab. 1 respectively. All other models, FCN (VGG and ResNet) and Deeplab v2 VGG were trained at 100% resolution, and Tab. 1 shows that perturbations generated from the multiscale and 100% resolutions of Deeplab v2 transfer the best. This supports the hypothesis that adversarial attacks produced at one scale are not as effective when evaluated at another since CNNs are not scale invariant (the network activations change considerably).

### 6.3. Multiscale networks and adversarial examples

The multiscale version of Deeplab v2 is the most robust to white-box attacks (Tab. 1, Fig. 2) as well as perturbations generated from single-scale networks. Moreover, attacks produced from it transfer the best to other networks as well, as shown by the bolded entries. This is probably because attacks generated from this model are produced from multiple input resolutions simultaneously. For the Iterative FGSM ll attack, only the perturbations from the multiscale version of Deeplab v2 transfer well to other networks, achieving a similar IoU ratio as a white-box attack. However, this is only the case when attacking a different scale of Deeplab. Whilst perturbations from multiscale Deeplab v2 transfer better on FCN than from single-scale inputs, they are still far from the efficacy of a white-box attack (which has an IoU ratio of 15.2% on FCN-VGG and 26.4% on FCN-ResNet).

Adversarial perturbations generated from multiscale inputs to FCN8 (which has only been trained at a single scale) behave in a similar way: FCN8 with multiscale inputs is more robust to white-box attacks, and its perturbations transfer better to other networks. This suggests that the observations seen in Tab. 1 are not properties of training the network, but rather the fact that CNNs are not scale invariant. Furthermore, an alternative to max-pooling the predictions at each scale is to average them. Average-pooling produces similar results to max-pooling. Details of these experiments, along with results using different attacks and $l_\infty$ norms ($\epsilon$ values), are presented in the supplementary.

### 6.4. Relation to other defenses

Our observations relate to the "random resizing" defense of [86] in concurrent work. Here, the input image is randomly resized and then classified. This defense exploits (but does not attribute its efficacy to) the fact that CNNs are not scale invariant and that adversarial examples were only generated at the original scale. Our findings suggest that this defense (which is very similar to the multiscale processing performed naturally by Deeplab v2) could be defeated by creating adversarial attacks from multiple scales, as done in this work, and this has indeed been verified [5, 85].

Table 1: Transferability of adversarial examples generated from different scales of Deeplab v2 (columns) and evaluated on different networks (rows). The underlined diagonals for each attack show white-box attacks. Off-diagonals, show transfer (black-box) attacks. The most effective one in bold, is typically from the multiscale version of Deeplab v2. The IoU ratio is reported.

| Network evaluated | FGSM ($\epsilon = 8$) | | | | Iterative FGSM ll ($\epsilon = 8$) | | | |
| --- | --- | --- | --- | --- | --- | --- | --- | --- |
| | 50% | 75% | 100% | Multiscale | 50% | 75% | 100% | Multiscale |
| Deeplab v2 50% scale (ResNet) | <u>37.3</u> | 70.5 | 84.8 | **60.3** | <u>18.0</u> | 92.0 | 96.9 | **20.0** |
| Deeplab v2 75% scale (ResNet) | 85.5 | <u>39.7</u> | 62.2 | **50.8** | 99.5 | <u>17.9</u> | 89.9 | **20.4** |
| Deeplab v2 100% scale (ResNet) | 93.6 | 57.9 | <u>37.7</u> | **37.2** | 100.0 | 79.0 | <u>15.5</u> | **16.8** |
| Deeplab v2 Multiscale (ResNet) | 83.7 | **57.6** | 62.3 | <u>53.1</u> | 99.6 | **90.2** | 91.9 | <u>21.5</u> |
| Deeplab v2 100% scale (VGG) | 94.3 | 70.6 | 66.9 | **66.5** | 98.9 | 88.4 | 86.3 | **80.9** |
| FCN8 (VGG) | 94.7 | 67.2 | 65.8 | **65.4** | 98.4 | 85.2 | 84.9 | **78.5** |
| FCN8 (ResNet) | 94.0 | 66.3 | 63.5 | **63.1** | 99.4 | 82.6 | 80.3 | **74.1** |

## 7. Image transformations and adversarial examples

In Sec. 6, we posited that adversarial examples are less malicious when processed at different scales since CNNs are not scale invariant. Scale changes are used in segmentation architectures to recognise objects at different resolutions, however, this is not the only commonly used image transformation. In this section, we consider a number of other common input transformations, and examine their effect on adversarial robustness of CNNs for semantic segmentation.

In the following, each transformation is applied to the input image before it is processed by the neural network and we examine how it affects the robustness to adversarial examples. Following on from Sec. 6, we use the Deeplab v2 MS network, which we found to be the most robust in Sec. 5, and consider the following four transformations (illustrated in Fig. 5) which are ubiquitous in computer vision and image processing:

**JPEG recompression.** The image is compressed using JPEG with a "quality" parameter drawn randomly between 50% and 100%. The image is then reconstructed and processed by the network.

**Gaussian blur.** The input image is blurred by a Gaussian filter with a bandwidth uniformly drawn from $[0, 2]$, which ensures that all objects in the image are still recognisable and can be segmented precisely.

**HSV jitter.** The image is converted to the HSV colour space (which is more perceptually similar than the RGB space). Next, each pixel is perturbed by a value drawn uniformly between $[-30, 30]$ and then converted back to RGB space for processing.

**Grayscale.** The input image is converted to grayscale by setting all three image channels to have the same value at each pixel. This was performed using a convex combination of each of the three RGB channels, with each of the co-efficients sampled from a flat Dirichlet distribution.

Note that none of the transformations affect the image spatial co-ordinates, which means that it is suitable for using with semantic segmentation models without any additional post-processing. These transformations, though quite disparate, all have a similar effect on adversarial robustness as described in the next subsection.

### 7.1. Robustness conferred by randomised input transformations

Figure 6 shows that each type of input transformation substantially increases the robustness of Deeplab v2 to the Iterative FGSM ll attack on the VOC dataset, with "JPEG recompression" and "Gaussian blur" providing substantial benefits. Converting the image to grayscale with random channel coefficients provides a smaller, but still sizeable, improvement. These findings are consistent and show little variance over 9 different trials, since each input transformation is randomised. The IoU of the transformed images at $\epsilon = 0$ (*i.e.* corresponding to no attack) is similar to the original image with the largest difference about 2%. Therefore, the network is more sensitive to input transformations on adversarial images than it is on clean ones.

These results, in addition to Tab. 6, show that as neural networks are not invariant to many classes of transformations of the input, their predictions on adversarial examples subject to these transformations change. Consequently, predictions on transformed adversarial inputs are different to the original adversarial example, and this typically results in the adversarial example becoming less malignant. These findings are consistent across a broad range of geometric and photometric transformations.

Dziugaite *et al.* [29] previously observed that JPEG recompression improved adversarial robustness for small $\epsilon$ values in the context of image classification. However, the authors hypothesised that a special property of the JPEG algorithm (*i.e.* mapping images back onto the manifold of natural images) was the reason it conferred additional ro-bustness. In contrast, our study of various different transformations suggest that JPEG recompression is just one instance of the numerous input transformations which neural networks are not invariant to. As a result, JPEG recompression, along with other image transformations, increases robustness to adversarial examples that were generated by attacks which did not take it into account.

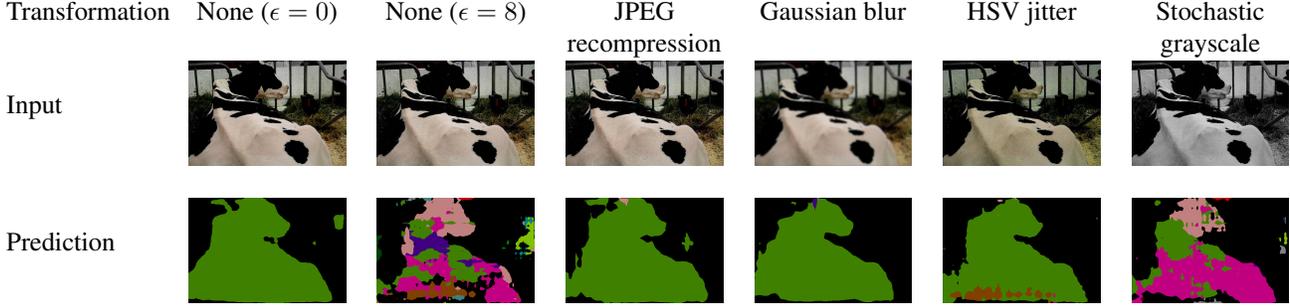

Figure 5: Input transformations of adversarial examples generated by Iterative FGSM ll (Eq. 6) significantly change the prediction of the Deeplab v2 network. These input transformations, however, barely change the output when they are applied to clean images. The $l_\infty$ norm of the perturbation, $\epsilon = 8$, is visible when looking carefully on screen.

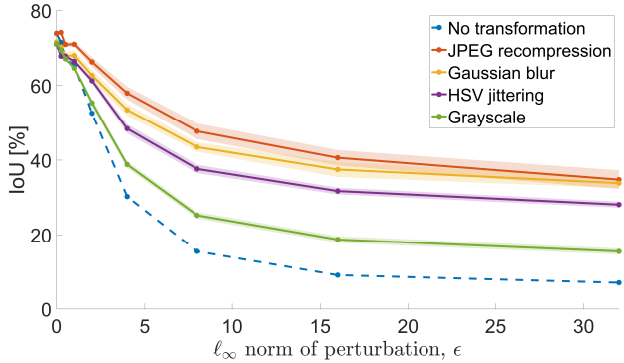

Figure 6: The adversarial examples originally generated by Iterative FGSM ll on Deeplab v2, are less malignant when the adversarial image is first pre-processed with a randomised transformation. The shaded regions correspond to two standard deviations computed from nine random trials of the randomised transformation.

### 7.2. Subverting randomised, non-differentiable input transformations

The results shown in Fig. 6 suggest that randomised input transformations serve as an effective defense to adversarial attacks. They significantly reduced the effectiveness of the Iterative FGSM ll attack, which has been the most powerful attack in our experiments, and the result for $\epsilon = 0$ also shows that this method has minimal performance penalties on clean inputs. This reasoning has been exploited by the concurrent work of [41], where the authors showed how several different input transformations increased the robustness of image classification models to adversarial attacks.

However, the results in Fig. 6 and [41] assume that knowledge of the defence mechanism (randomised input transformations in this case) is not exploited in generating the adversarial attack. This methodology goes against Kerckhoffs' principle [50] – the basis of modern cryptographic systems – which states that a system should be secure if everything about it barring the key is public knowledge.

Consequently, to confirm if randomised input transformations really confer adversarial robustness, we modify the Iterative FGSM ll update (Eq. 6) to compute the expected gradient over the distribution of transformations which could be applied at inference time,

$$\mathbf{x}_{t+1}^{adv} = \text{clip}\Big(\mathbf{x}_t^{adv} - \alpha \cdot \text{sign}(\mathbb{E}_{t \sim \mathcal{T}} \nabla_{\mathbf{x}_t^{adv}} L(f(t(\mathbf{x}_t^{adv}); \theta), y_{ll}), \epsilon)\Big), \quad (7)$$

where $\mathcal{T}$ is the distribution of transformation functions $t$. This method uses the fact that $\nabla_\mathbf{x} \mathbb{E}_{t \sim \mathcal{T}} f(t(x)) = \mathbb{E}_{t \sim \mathcal{T}} \nabla_\mathbf{x} f(t(x))$. It has also been used by a concurrent work [5] to estimate the gradient of neural networks with randomised non-differentiable adversarial defences [86]. This variant of the FGSM attack corresponds to sampling from the distribution of transformations, computing the loss and gradient of the image with respect to the loss, and averaging this gradient over many samples before performing the update.

Note that some transformations, such as JPEG recompression, are not differentiable. In this case, we use the straight-through estimator [9] which assumes, when computing the gradient using backpropagation, that the transformation is the identity function.

Figure 7 shows the results of the Expectation over Transformations (EOT) attack (Eq. 7) on the Deeplab v2 model on the Pascal VOC dataset, with the expectation computed over 10 samples. The randomised JPEG and Gaussian blur input transformations increase the robustness of the model marginally, whilst jittering pixel values in the HSV space and grayscale conversion provide no additional robustness.

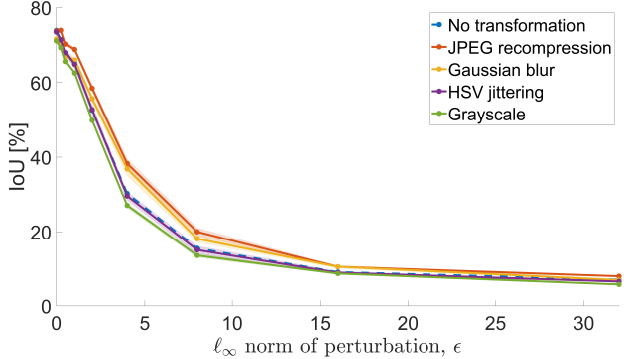

Figure 7: The randomised input transformations no longer increase the robustness of the network when the expected gradient over the distribution of the transformation functions is used in the Iterative FGSM ll attack. The shaded regions correspond to two standard deviations computed from nine random trials of the randomised transformation. The dashed blue line shows the original Iterative FGSM ll attack on non-transformed images.

The final IoU is similar to the original model that did not use randomised input transformations and was attacked with the standard Iterative FGSM ll attack. To our knowledge, we are the first who show that neural networks can easily be attacked with both non-differentiable and randomised input transformations. However, we point out that [5] have attacked numerous recent defenses, some which were non-differentiable or randomised, but not both.

### 7.3. Transferability of input transformations

The previous two parts have shown that using input transformations reduces the malignancy of an adversarial perturbation (Sec. 7). Our second observation however showed that whenever we exploit knowledge about the input transformation during attack generation, the perturbation can become as malignant as the attack on the image with no input transformation (Sec. 7.2).

In this section, we examine the transferability of the perturbations generated from different transformations as described in Sec. 7.2. For example, we consider the efficacy of a perturbation created using the "JPEG recompression" transformation when the network's input is pre-processed with "Gaussian blur" instead. This has important implications on the robustness and security of neural networks; if the perturbations do not transfer across different input transformations, it would suggest that a "security-through-obscurity" approach could be used, as a defender could secure their system by ensuring that the attacker does not know the input transformations they are using It also has implications on our ability to produce malicious physical adversarial examples [54, 77], as physical objects in the real world can be viewed from a diverse range of illumination conditions, camera viewpoints and other transformations of an original canonical view.

Table 2 and Fig. 8 show our results when the adversarial perturbation generated using one distribution of transformations is applied on a network using another randomised transformation as pre-processing. Table 2 shows the absolute IoU (to account for the fact that input transformations cause slight changes on the accuracy of clean inputs) for $\epsilon = 8$, which is when the adversarial perturbations become conspicuous to the human eye, whilst Fig. 8 summarises the results for all $\epsilon$ values. Perturbations generated to target "JPEG recompression" or "Gaussian blur" input pre-processing (the two transformations which confer the most robustness to standard attacks generated without transformations (Fig. 6)), show poor transferability when the "Grayscale" or "HSV jitter" input transformation is used instead. In contrast, perturbations generated to target the "Grayscale" input transformation transfer the best to the other input transformations that we have considered in our experiments. Additionally, the last row of Tab. 2 shows that when no input transformation is used at inference time, attacks generated to target a particular input transformation are more effective with the exception of the "Grayscale" transformation. This corresponds with our results in Sec. 6 where adversarial attacks generated at multiple scales transferred better to other models.

There are clearly a myriad of input transformations that could be performed as input pre-processing to a neural network, of which we have considered only a handful. Nevertheless, it is evident that targeting some input transformations (such as grayscale conversion) appears to produce perturbations that are more transferable to other input transformations in comparison to others (JPEG recompression). This raises an important research question about why including certain input transformations into the attack generation process transfer better to other input transformations. It also suggests another critical and open question, whether it is possible to produce adversarial perturbations that are malignant across all input transformations without modelling all of these transformations explicitly when generating the attack.

### 7.4. Relation to concurrent work

Our findings corroborate with concurrent work discussing producing physical adversarial attacks. Lu *et al*. [61] created adversarial traffic signs by capturing images of road signs from 0.5m and 1.5m away, generating attacks from these images on a computer, and then printing out the adversarial image onto paper. Whilst the printed image taken from 0.5m away fooled an object detector viewing the adversarial image from 0.5m, it did not when viewed from 1.5m and vice versa. This result is corroborated by Tab. 1 which shows that adversarial examples transfer poorly across different scales. Subsequent work [6, 32] has shown that it is possible to construct adversarial examples that are malignant across multiple different scales by incorporating scale changes into the attack generation process. This is again supported by our results in Tab. 1, and Sec. 7.2 which also show this effect for a number of other input transformations. When producing physical adversarial attacks, it is difficult to model all the transformations that the original image could be subject to, and as reflected by Sec. 7.3, adversarial examples generated to target a particular transformation do not always transfer well to other input transformations. This may explain why the adversarial traffic signs generated by [32] have not been able to fool the detectors subsequently evaluated by Lu *et al*. [62]. Our observation that input transformations that were not explicitly modelled in the attack generation process mitigate the effectiveness of adversarial attacks also suggest that future work on physical adversarial attacks requires much more robust evaluation than initial work in this area [54, 32, 61, 12]. This is to ascertain whether the proposed attacks are still effective in the diverse environmental conditions that images of the adversarial object may be acquired from.

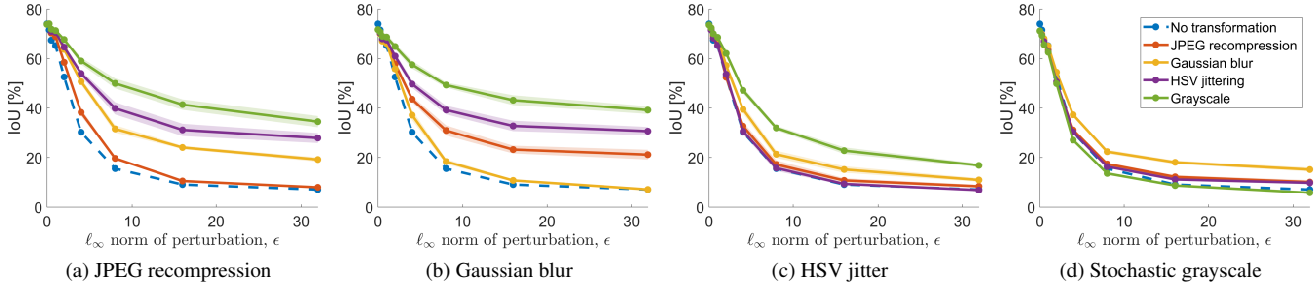

(a) JPEG recompression  (b) Gaussian blur  (c) HSV jitter  (d) Stochastic grayscale

Figure 8: The effectiveness of adversarial examples generated with one distribution of input transformations, and evaluated with another. The title of each graph shows the input transformation the adversarial examples were generated with. Each graph is effectively a column of Tab. 2 for multiple $\epsilon$ values. The dotted blue line shows the Iterative FGSM ll attack when input transformations are not used at either inference or attack generation time.

Table 2: Transferability of adversarial attacks generated with different input transformation distributions. The left column indicates the distribution of transformations (as described in Sec. 7) that was used at inference time, and the other columns show the input transformations used when generating the attack. This table shows the mean absolute IoU scores of the Deeplab v2 network on the VOC dataset for the Iterative FGSM ll attack with $\epsilon = 8$. The diagonals show "white-box" entries where the input transformation distribution used at inference time is used to generate the attack as well. The bold entries off the diagonals show the strongest attack when a different transformation distribution is used at inference time.

| Input transformation at inference time | Input transformation to generate attack | | | | |
|---|---|---|---|---|---|
| | JPEG recompression | Gaussian blur | HSV jittering | Stochastic grayscale | None |
| JPEG recompression | <u>19.7</u> | 30.9 | 17.2 | **17.4** | 47.7 |
| Gaussian blur | 31.6 | <u>18.4</u> | **21.3** | 22.4 | 43.5 |
| HSV jitter | 39.9 | 39.2 | <u>15.7</u> | **16.3** | 33.5 |
| Stochastic grayscale | 50.0 | 49.3 | 32.0 | <u>13.6</u> | **25.2** |
| None | **11.6** | 14.4 | 12.0 | 24.4 | <u>15.5</u> |

Our study of the effect of input transformations on adversarial robustness also emphasises the importance of incorporating knowledge of the proposed adversarial defence into the attack used to validate it (Kerckhoff's principle [50]). This is not the case for many recently proposed defenses [41, 86, 13, 56] which have all subsequently been defeated [17, 5, 85, 4].

## 8. Effect of CRFs on Adversarial Robustness

Conditional Random Fields (CRFs) are commonly used in semantic segmentation to enforce structural constraints [3]. The most common formulation is DenseCRF [52], which encourages nearby (in terms of position or appearance) pixels to take on the same label and hence prefers smooth labelling. This is done by a pairwise potential function, defined between each pair of pixels, which takes the form of a weighted sum of a bilateral and Gaussian filter.

Intuitively, one may observe that adversarial perturbations typically appear as a high-frequency noise, and thus the pairwise terms of DenseCRF which act as a low-pass filter, may provide resistance to adversarial examples. To verify this hypothesis, we consider CRF-RNN [93]. This approach formulates mean-field inference of DenseCRF as an RNN which is appended to the FCN8s network [60], enabling end-to-end training.

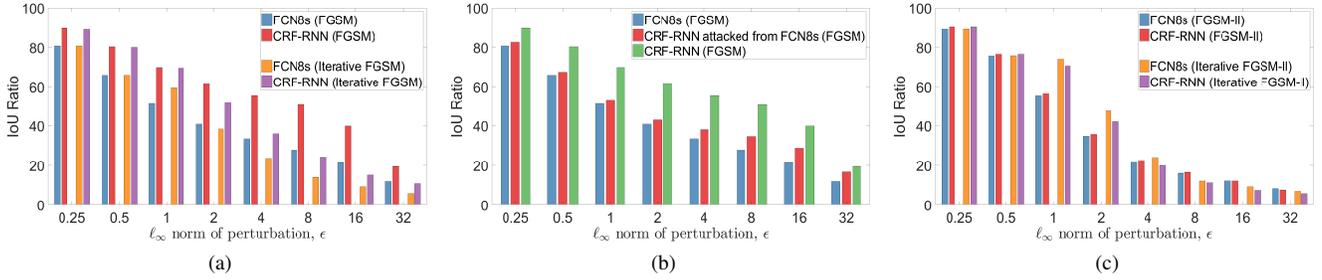

Figure 9: (a) On untargetted attacks on Pascal VOC, CRF-RNN is noticeably more robust than FCN8s. (b) CRF-RNN is more vulnerable to black-box attacks from FCN8, due to its "gradient masking" effect which results in ineffective white-box attacks. (c) However, the CRF does not "mask" the gradient for targeted attacks and it is no more robust than FCN8s.

### 8.1. CRFs confer robustness to untargeted attacks

Fig. 9a shows that CRF-RNN is markedly more robust than FCN8s to the untargeted FGSM and Iterative FGSM attacks. To verify the hypothesis that the smoothing effect of the pairwise terms increases the robustness to adversarial attacks, we evaluated various values of the bandwidth hyperparameters defining the pairwise potentials (not learned; in Fig. 9a, we used the values of the public model).

Higher bandwidth values (increasing smoothness) do not actually lead to greater robustness. Instead, we observed a correlation between the final confidence of the predictions (from different hyperparameter settings) and robustness to adversarial examples. We measured confidence according to the probability of the highest-scoring label at each pixel, as well as the entropy of the marginal distribution over all labels at each pixel. The mean confidence and entropy for CRF-RNN (with original hyperparameters) is 99.1% and 0.025 nats respectively, whilst it is 95.2% and 0.13 nats for FCN8s (additional details in supplementary). The fact that mean-field inference tends to produce overconfident predictions has also been noted previously by [68] and [15].

More confident predictions lead to a smaller loss, making attacks which use the gradient of the loss with respect to the input less effective. The "Defensive Distillation" approach of [73] made use of a similar fact by increasing the confidence of the model's predictions, resulting in gradients of smaller norm. The key difference is that CRFs increase the confidence as a by-product of a technique generally used to improve accuracy on numerous pixel-wise labelling tasks, while the effect of [73] on accuracy is unknown, as it was only tested on the saturated MNIST and CIFAR10 datasets.

### 8.2. Circumventing the CRF

Although CRFs are more resistant to untargeted attacks, they can still be subverted in two ways. CRF-RNN is effectively FCN8s with an appended mean-field layer. Fig. 9b shows, that adversarial examples generated via FGSM from FCN8s ("unary" potentials) are more effective on CRF-RNN than attacks from the output layer of CRF-RNN.

Also, targeted attacks with FGSM ll and Iterative FGSM ll are more effective since the label used to compute the loss for generating the adversarial example is not the network's (highly confident) prediction but rather the least likely label. Consequently, the loss is high and there is a strong gradient signal from which to compute the adversarial example. Fig. 9c shows that CRF-RNN and FCN8s barely differ in their adversarial robustness to targeted attacks.

Finally, Fig. 10 shows that the same observations hold on the DeepLab v2 network, where the DenseCRF model is used as post-processing, and is not part of the neural network. This confirms that end-to-end training of the CRF, as done in CRF-RNN [93], does not influence adversarial robustness.

### 8.3. Discussion

The smoothing effect of CRFs, perhaps counterintuitively, has no impact on the adversarial robustness of a DNN. However, mean-field inference produces confident marginals, making untargeted attacks less effective since they rely on the gradient of the final loss with respect to the prediction. Black-box attacks generated from models without a CRF transfer well to networks with a CRF, and are actually more effective. This is the case for both CRFs trained end-to-end [93] and used as post-processing [22], as shown in the supplementary. Finally, CRFs confer no robustness to untargeted attacks. Our investigation of the CRF also underlines the importance of testing thoroughly with black-box attacks and multiple attack algorithms, which is not the case for numerous proposed defenses [24, 37, 38, 73].

## 9. Conclusion

We have presented what to our knowledge is the first rigorous evaluation of the robustness of semantic segmentation models to adversarial attacks. We believe our main observations will facilitate future efforts to understand and defend

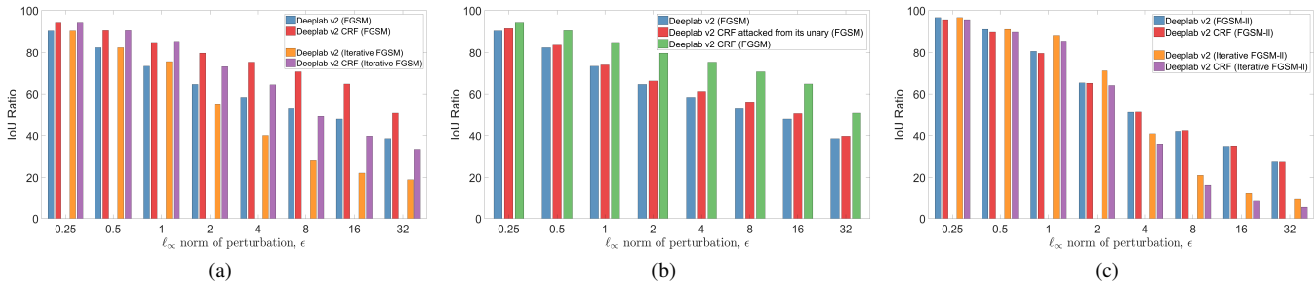

Figure 10: Similar trends are observed for Deeplab v2, which uses the DenseCRF model as post-processing, as CRF-RNN (Fig. 9) which integrates the CRF as part of the deep network. (a) On untargetted attacks, Deeplab v2 with a CRF is noticably more robust than just the Deeplab v2 network. (b) Attacks created from the base Deeplab v2 network using FGSM are more effective than those created from Deeplab v2 with CRF. This is due to the "gradient masking" effect of mean-field inference of CRFs. (c) However, the CRF does not "mask" the gradient for targeted attacks. As a result, Deeplab v2 with a CRF is no more robust than just the Deeplab v2 network.

against these attacks without compromising accuracy:

Networks with *residual connections* are inherently more robust than chain-like networks. This extends to the case of models with very few parameters, contrary to the prior observations of [55, 63]. *Multiscale* processing makes CNNs more robust since adversarial inputs are not as malignant when processed at a different scale from which they were generated at, probably as CNNs are not invariant to scale. Using other *input transformations* that CNNs are not invariant make them markedly more robust to transformed adversarial examples but only when the attack generation does not take knowledge of these input transformations into account. This holds even when the input transformations are randomised, however, when this knowledge is taken into account during attack generation, only marginal improvements in robustness are observed. The fact that adversarial attacks generated to target particular input transformation do not always transfer well to other input transformations also suggests that producing physical adversarial attacks in varying environmental conditions is difficult.

*Mean-field inference for Dense CRFs*, which increases the confidence of predictions confers robustness to untargeted attacks, as it naturally performs "gradient masking" [71, 73]. There are no robustness benefits from the smoothness priors enforced by the DenseCRF model.

In the shorter term, our observations suggest that networks such as Deeplab v2, which is based on ResNet and performs multiscale processing, should be preferred in safety-critical applications due to their inherent robustness. As the most accurate network on clean inputs is not necessarily the most robust network, we recommend evaluating robustness to a variety of adversarial attacks as done in this paper to find the best combination of accuracy and robustness before deploying models in practice. We also emphasize that it is crucial to evaluate proposed defenses judiciously, *e.g.* using the white-box attacks which exploit knowledge of the proposed defense to assess the real efficacy of such a defense.

Adversarial attacks are arguably the greatest challenge affecting DNNs. The recent interest of our field into this phenomenon is only the start of an important longer-term effort, and we should also study the influence of other factors such as training regimes and attacks tailored to evaluation metrics. In this paper, we have made numerous observations and raised questions that will aid future work in understanding adversarial examples and developing more effective defenses.

## Acknowledgments

This work was supported by the EPSRC, Clarendon Fund, ERC grant ERC-2012-AdG 321162-HELIOS, EPSRC grant Seebibyte EP/M013774/1 and EPSRC/MURI grant EP/N019474/1.

# Appendix

This supplementary material details the DNN models we analysed, and experiments we omitted from the main paper since they follow similar trends. Section A provides further details about the experimental set-up, including the various DNNs used in the experiments. Section B shows qualitative examples of the adversarial attacks we studied. Section C presents further experimental results about "The robustness of different networks" (Sec. 5 of the main paper). Similarly, Section D shows more experimental results about "Multi-scale Processing and Transferability of Adversarial Examples" (Sec. 6 of the main paper). Finally, Section E presents further experimental results on the "Effect of CRFs on Adversarial Robustness" (Sec. 8 of the main paper).

## A. Experimental setup

This section details the DNN models, additional information about the Cityscapes dataset and the software and hardware used in the experiments.

### A.1. Software and hardware setup

We use the Caffe [48] deep learning framework for all experiments, since most publicly available segmentation models are implemented using this library. Our experiments are performed on either a Nvidia M40 or P100 GPU which have 12GB and 16GB of memory respectively.

### A.2. Description of models

We detail each model in this section. Tab. 3 shows the performance of publicly available models on the Pascal VOC validation set. Tab. 4 compares the Intersection over Union (IoU) obtained by models that we have retrained compared to the original author's performance where available. Tab. 5 shows the performance of publicly available models on the Cityscapes validation set. Finally, Tab. 6 lists the number of parameters in each of the models.

**FCN8s [60].** We retrained the FCN8s (VGG) network on Pascal VOC with additional annotations from SBD [42] and MS-COCO [58]. The publicly available model of FCN8s is not trained with MS-COCO, which is why we retrained it ourselves. As shown in Tab. 4, we obtain an IoU of 68.7% on the VOC validation set, whilst the original authors who did not train on MS-COCO obtained 65.5% [78].

For the Cityscapes dataset, we used the publicly available VGG model[2] from [79].

We trained FCN8s with a ResNet-101 backbone on Pascal VOC since no publicly available model was available. As shown in Tab. 4, the IoU on clean inputs of this version is close to the VGG version. We are not aware of any other published work to compare this number to.

---

[2] https://github.com/shelhamer/clockwork-fcn
MD5 checksum of Caffe model: fcae4fdc759f9f461fffc7cc3baa96c6

Table 3: Networks with public models, evaluated on the VOC validation set

| Model Name | IoU [%] |
| --- | --- |
| CRF-RNN [93] | 72.8 |
| Dilated Frontend [90] | 67.1 |
| Dilated Context [90] | 70.4 |
| SegNet [7] | 43.0 |

**Deeplab v2 [22].** We cannot use the publicly released models for the Pascal VOC dataset, since they have been trained on the entire validation set as well. Hence, we use the authors' publicly released training code[3] to retrain their networks without the VOC validation set.

We retrained the Deeplab v2 network with ResNet-101 and VGG backbones on Pascal VOC, achieving similar performance to the original authors as shown in Tab. 4. Note that the authors [22] reported results from ablation experiments on the VOC validation set, which we compare to in Tab. 4. However, these models have never been released.

For CRF post-processing, we used the hyperparameters used by the original authors. As the weights of our trained model are different to the authors, it is possible that different CRF hyperparameters that obtain a higher IoU on the validation set exist.

**PSPNet [92].** We used the publicly available model[4] for our experiments on Cityscapes. As the public VOC model has been trained on the entire validation set, we cannot use it for our experiments. Consequently, we retrained this model ourselves achieving comparable results to the original authors (Tab. 4). We followed the training procedure described in the original paper where possible. However, the original authors trained the model using 16 GPUs allowing an effective batch size of 16. Due to our limited computational resources, we could only train on a single GPU using a batch size of 1. The large batch size enabled the original authors to compute better batch statistics for batch normalisation. When using a batch size of 1, the variance in the batch statistics is too high to perform batch normalisation. As a result, we "froze" our batch normalisation layers, and used the batch statistics (mean and variance) of the ImageNet-pretrained ResNet-101 model. This is common practice in training semantic segmentation [22] and ob-ject detection [46] networks where batch sizes are typically small.

---

[3] https://bitbucket.org/aquariusjay/deeplab-public-ver2.git
[4] https://github.com/hszhao/PSPNet
MD5 checksum of Caffe model: 29bbdf0ce4d2a6546ed473656db1d6e2

Table 4: Retrained models on VOC validation set. Details about FCN8, Deeplab v2 and PSPNet can be found in Sec. A.2.

| Model Name | IoU [%] | IoU of authors [%] |
|---|---|---|
| FCN8s (VGG) [60] | 68.7 | – |
| FCN8s (ResNet) [60] | 68.8 | – |
| Deeplab v2 ASPP (VGG) [22] | 66.9 | 68.9 |
| Deeplab v2 ASPP (ResNet) [22] | 73.3 | – |
| Deeplab v2 Multiscale ASPP (ResNet) [22] | 73.9 | 76.3 |
| Deeplab v2 Multiscale ASPP (ResNet) + CRF post-processing [22] | 74.9 | 77.7 |
| PSPNet [92] | 75.9 | – |
| PSPNet [92] (test set) | 79.0 | 85.4 |

Table 5: Networks with public models on Cityscapes validation set. We have reported the IoU at $1024 \times 512$ inputs, as well as the original $2048 \times 1024$ if the network was trained using full-resolution crops.

| Model name | IoU at $1024 \times 512$ | IoU at $2048 \times 1024$ |
|---|---|---|
| E-Net [74] | 53.4 | – |
| ICNet [91] | 56.5 | 67.2 |
| FCN8s (VGG) [79] | 62.1 | 66.4 |
| Dilated Frontend [90] | 59.0 | 64.6 |
| Dilated Context [90] | 62.3 | 68.6 |
| PSPNet [92] | 74.4 | 79.7 |

Table 6: The number of parameters in each of the DNN models evaluated in this paper. As all the networks are stored as 32-bit/4-byte floating point numbers, we reported the number of parameters in megabytes (MB).

| Model Name | Dataset | Number of parameters (MB) |
|---|---|---|
| E-Net | Cityscapes | 1.5 |
| ICNet | Cityscapes | 30.1 |
| PSPNet (ResNet-101) | Cityscapes | 260.2 |
| Dilated Frontend (VGG) | Cityscapes | 512.4 |
| FCN8s (VGG) | Cityscapes | 512.5 |
| Dilated Context (VGG) | Cityscapes | 512.6 |
| Segnet (VGG) | Pascal | 112.4 |
| Deeplab v2 (VGG) | Pascal | 144.5 |
| FCN8s (ResNet-101) | Pascal | 162.9 |
| Deeplab v2 (ResNet-101) | Pascal | 168.4 |
| PSPNet (ResNet-101) | Pascal | 272.7 |
| Dilated Frontend (VGG) | Pascal | 512.4 |
| FCN8s (VGG) | Pascal | 513.0 |
| CRF-RNN (VGG) | Pascal | 513.0 |
| Dilated Context (VGG) | Pascal | 538.4 |

As shown in Tab. 4, our reimplementation of PSPNet on VOC achieves comparable results to the original authors, even though it has been trained on 1449 fewer images (the VOC validation set). We compared our implementation to the authors on the held-out test set (evaluation is performed on an online server) as the performance on the validation set is not reported in the original paper.

**CRF-RNN [93].** We used the publicly available model for Pascal VOC (trained on MS-COCO)[5].

**DilatedNet [90].** We used the public Pascal VOC and Cityscapes models[6].

**ICNet [91].** We used the public Cityscapes model[7].

**E-Net [74].** We used the public Cityscapes model[8].

**SegNet [7].** We used the public Pascal VOC model[9].

### A.3. Cityscapes dataset

Tab. 5 shows the performance of various publicly available models on the Cityscapes validation set consisting of 500 images. Cityscapes images are captured at a high resolution of $2048 \times 1024$, which is too large to fit into GPU memory for most networks. With the exception of E-Net [74] (which is trained on half-resolution images), the other networks we evaluated are trained on smaller crops of full-resolution images. Thereafter, at test time, authors use different tiling strategies [90, 92] to process parts of an image at full resolution before combining the partial results. To make a fairer comparison between models, we process all images at half-resolution so that tiling is not required. In Tab. 5, we show the IoU at the resolution we tested on, $1024 \times 512$. And if the model was also trained on full resolution crops, we also include the IoU of the network on full resolution inputs.

---

[5] https://github.com/torrvision/crfasrnn
MD5 checksum of Caffe model: bc4926ad00ecc9a1c627db82377ecf56
[6] https://github.com/fyu/dilation.
MD5 checksum for Pascal VOC: 7a44221dbc2611529bff32029ad1f6e2
MD5 checksum for Cityscapes: 0de4d78b5f9692f2aba5e7ed88f93ccb
[7] https://github.com/hszhao/ICNet
MD5 checksum of Caffe model: c7038630c4b6c869afaaadd811bdb539
[8] https://github.com/TimoSaemann/ENet
MD5 checksum of Caffe model: d9aabd630cf6bc29c48ea55a86124e14
[9] https://github.com/alexgkendall/
SegNet-Tutorial/blob/master/Example_Models/
segnet_model_zoo.md
MD5 checksum of Caffemodel: 6e01077e3cda996f95b2a82ea4641a4c

## B. Qualitative results

Figure 11 visualises adversarial perturbations of varying $\ell_\infty$ norms, showing how the perturbations only become visible to the naked eye when the $l_\infty$ of the perturbation, $\epsilon$, is 8 (when viewed on screen). Figure 12 shows the results of the four adversarial attacks considered in this paper when applied on the same image from the Pascal VOC dataset on the Deeplab v2 network. Finally, Fig. 13 compares the outputs of different networks to the Iterative FGSM ll attack for varying values of $\epsilon$ on the Cityscapes dataset.

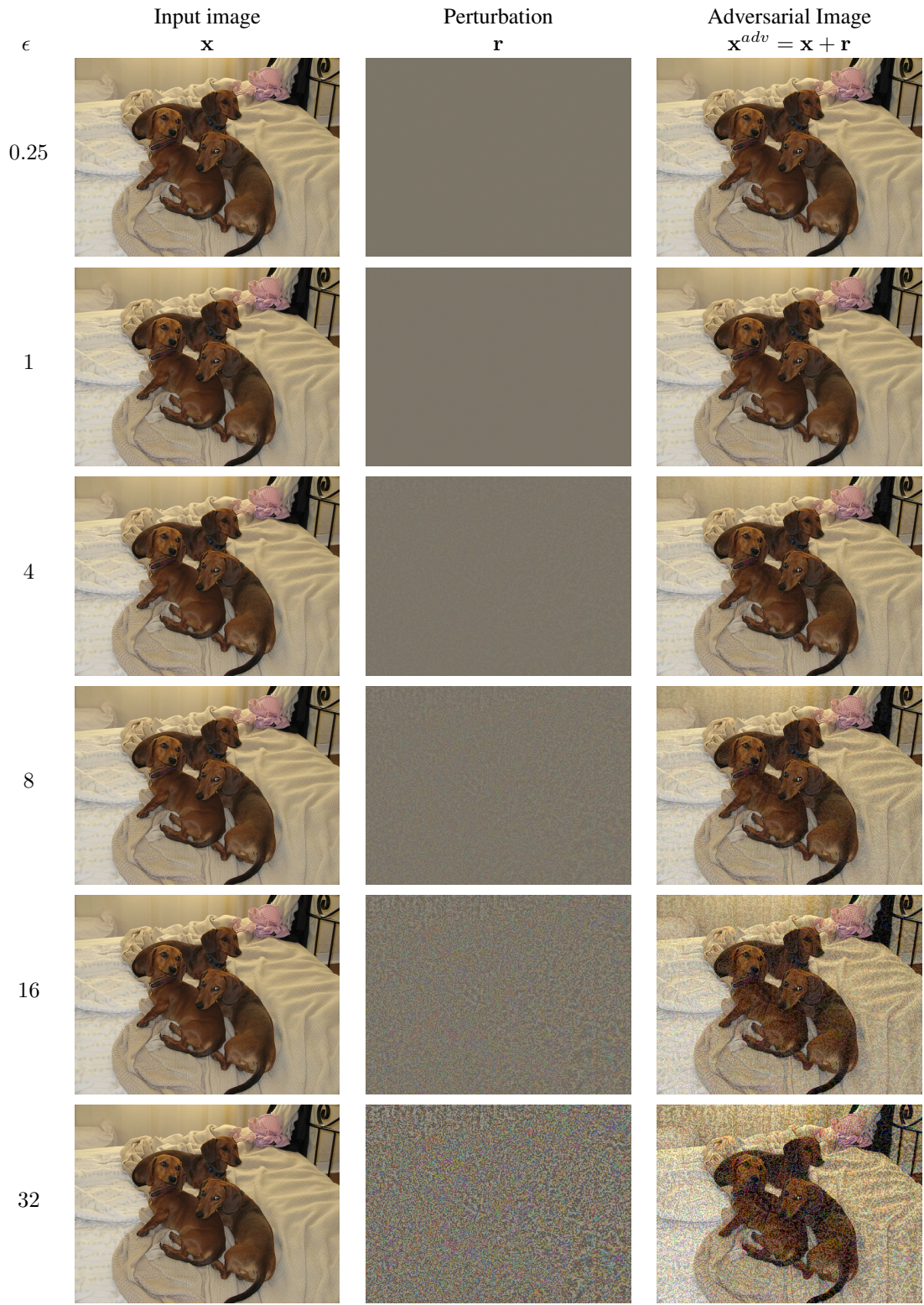

Figure 11: A visualisation of adversarial perturbations of varying $\ell_\infty$ norms. The perturbation, in the middle column, when added to the input, produces the adversarial example that fools neural networks. Note that the mean RGB value (of the Pascal VOC dataset) is already added to the perturbation, resulting in the grey background. This is required for visualisation as the perturbation can be negative, and RGB images are stored as positive integers $\in [0, 255]$. For $\epsilon = 0.25$, the adversarial image and input image are actually identical if rounded to integers (as RGB images are typically represented). Nevertheless, perturbations of this norm have fooled every neural network studied in this paper. Perturbations become noticeable when viewed on screen at around $\epsilon = 8$. In this figure, perturbations were created using FGSM on Deeplab v2.

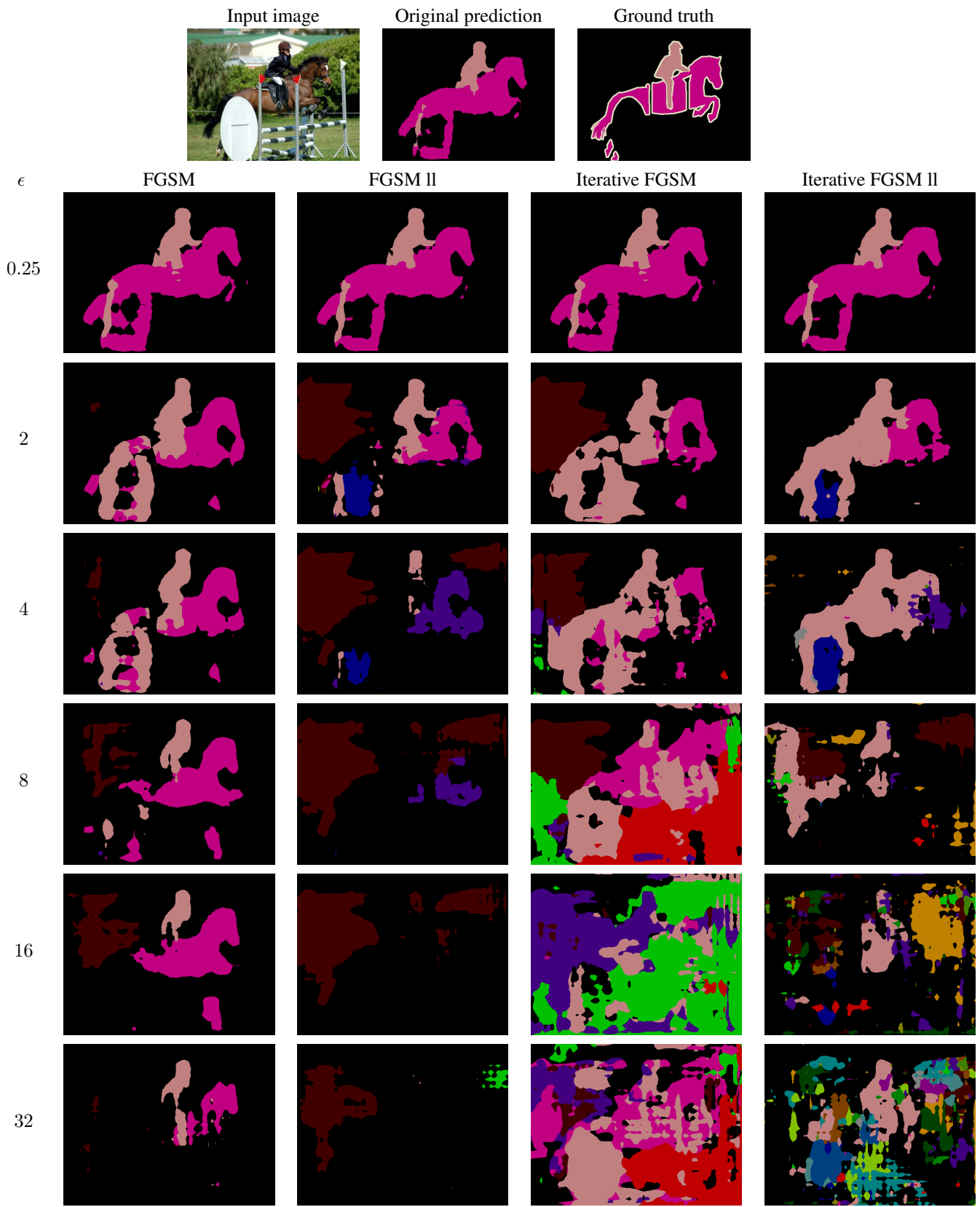

Figure 12: A comparison of different adversarial attacks on the Deeplab v2 Mulitscale ASPP network [22], on a common image from Pascal VOC. As expected, iterative attacks (last two columns) are more effective than single-step ones (first two columns). Higher $l_\infty$ norms of the perturbation, $\epsilon$, also degrade the network's prediction more.

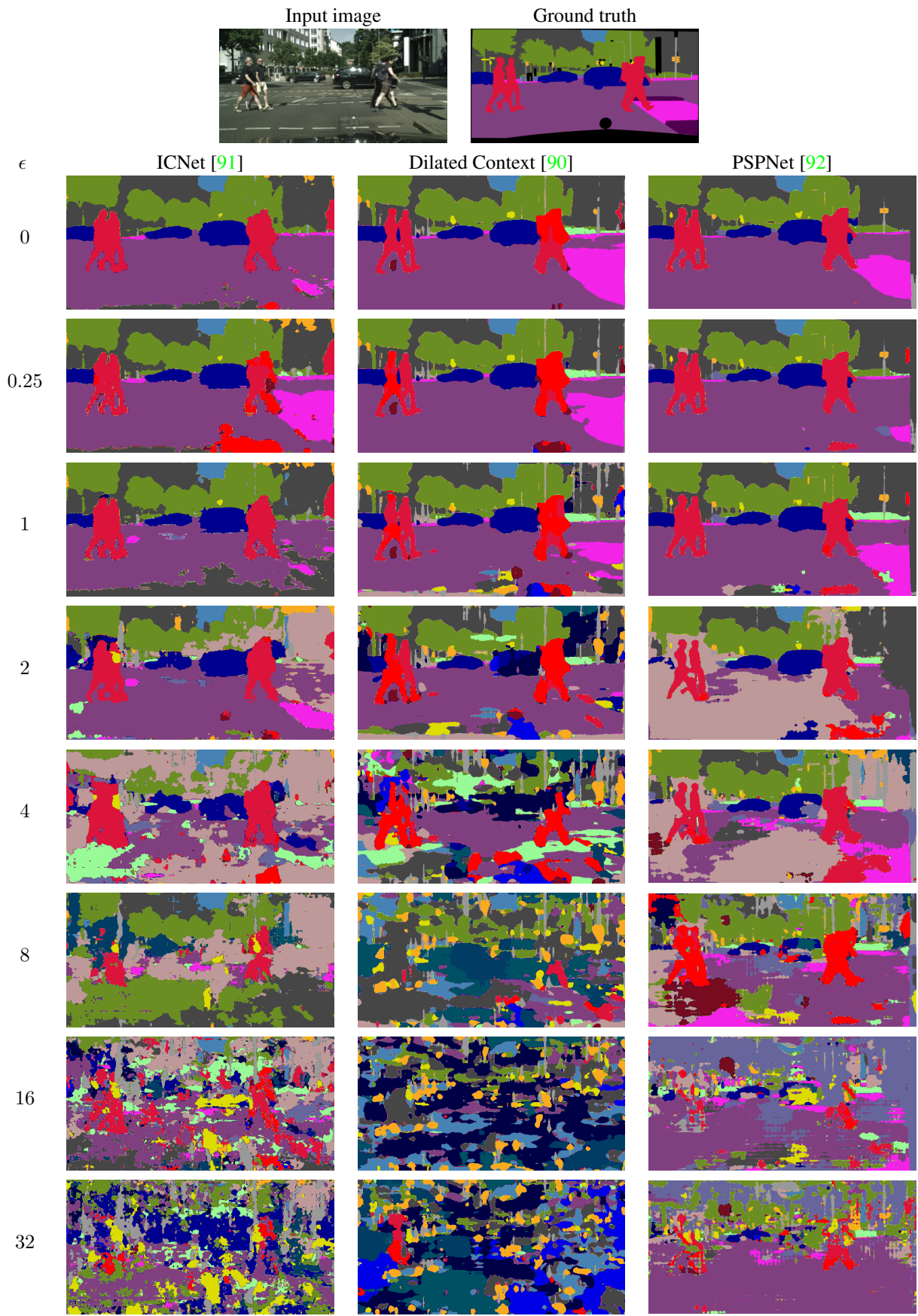

Figure 13: Comparison of ICNet, Dilated Context and PSPNet when attacked by Iterative FGSM ll, for different values of the $l_\infty$ norm, $\epsilon$. Note how each network is affected differently, with PSPNet the most robust. $\epsilon = 0$ is the original prediction of the network, since no perturbation is added here.

## C. Robustness of Different Architectures

The main paper presented results using the FGSM and Iterative FGSM ll attacks for both Pascal VOC and Cityscapes datasets. In this section, we present results for the targeted, single-step FGSM ll and untargeted Iterative FGSM attacks as well. Furthermore, we also include the Absolute IoU scores for each attack for different $l_\infty$ perturbations.

### C.1. Results of other attacks

Figures 14 and 15 show results of the FGSM ll and Iterative FGSM attacks on the VOC and Cityscapes datasets respectively. Our primary observations from the main paper are mostly consistent on these attacks as well:

- ResNet based networks are more robust than models based on VGG.

- DilatedNet [90] without its "Context" module is typically more robust than the full, more accurate network.

- E-Net and ICNet show similar robustness to DilatedNet on the Cityscapes dataset. It is only for the FGSM ll attack for $\epsilon \geq 4$ that DilatedNet is robust than both of these lightweight networks.

- Single-step attacks (FGSM ll) are particularly effective on Cityscapes at high $\epsilon$ values. They are more effective at fooling networks than iterative methods as well. This was unexpected, and not observed on Pascal VOC.

- PSPNet, which achieves the highest IoU on clean inputs, is typically not the most robust network on Pascal VOC.

### C.2. Result tables of Absolute IoU

In contrast to the main paper that showed the IoU Ratio for various attacks, Tables 7 through 10 show the absolute IoU for different models for each of the FGSM, FGSM ll, Iterative FGSM and Iterative FGSM ll attacks on the Pascal VOC dataset. Additionally, Tables 11 through 14 show the absolute IoU for different models on the Cityscapes dataset.

Note that PSPNet, which achieves the highest IoU on clean inputs, does not usually achieve the highest absolute IoU when attacked on the Pascal VOC dataset. When considering 4 adversarial attacks, and 8 $\epsilon$ values, PSPNet achieves the highest absolute IoU in only 2 out of 32 cases. Moreover, it never achieves the highest absolute IoU for imperceptible perturbations ($0 < \epsilon \leq 4$).

Additionally, the highest absolute IoU for any $\epsilon$ value is always from a ResNet-based model (Deeplab v2, FCN8s (ResNet) or PSPNet) on the Pascal VOC dataset. On Cityscapes, FCN8s (VGG) is sometimes the most robust network at high $\epsilon$ values. However, the performance of all the networks is severely degraded at this point.

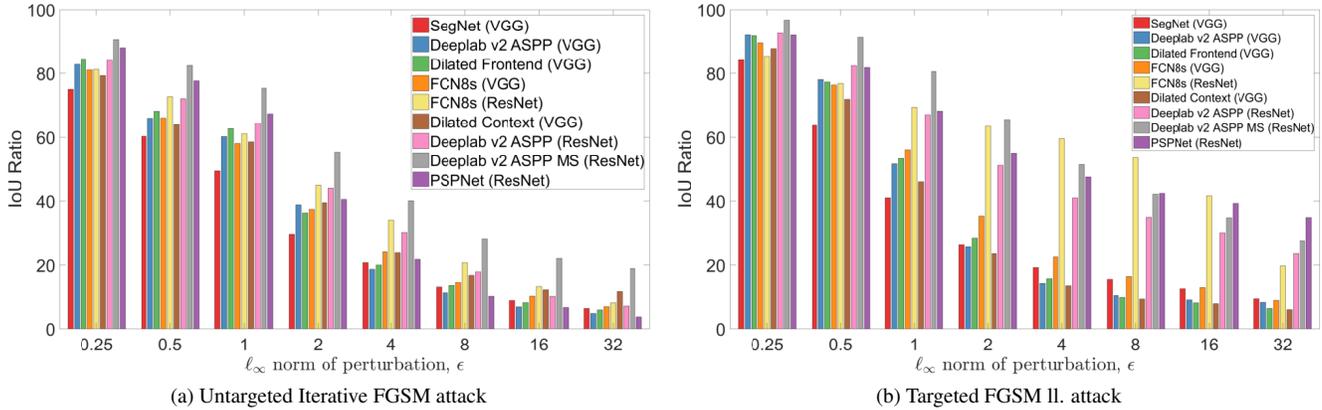

Figure 14: Adversarial robustness of state-of-the-art models on the Pascal VOC dataset. As with the FGSM and Iterative FGSM ll attacks in the main paper, models based on the ResNet backbone are more robust. Deeplab v2 is generally the most robust network, except on the Targeted FGSM attack for $\epsilon \geq 4$. The Iterative FGSM attack is also more effective at fooling the networks than the single-step Targeted FGSM attack, as shown by the lower IoU ratios.

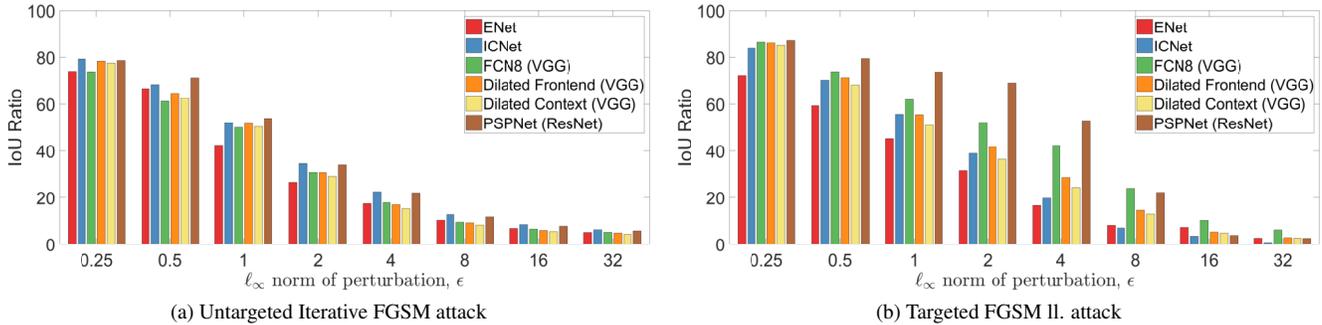

Figure 15: Adversarial robustness of state-of-the-art models on the Cityscapes dataset. As with the FGSM and Iterative FGSM ll attacks in the main paper, PSPNet is typically the most robust. Once again, DilatedNet without its "Context" module is slightly more robust than the full, more accurate network. The single-step FGSM ll attack is also more effective at higher $\epsilon$ values than the Iterative FGSM attack. This is unexpected, but was also observed in the main paper between the FGSM and Iterative FGSM ll attacks.

Table 7: The absolute IoU on the *Pascal VOC* dataset for various models when attacked with *FGSM*. This is evaluated for eight different values of the $\ell_\infty$ norm of the perturbation, $\epsilon$. $\epsilon = 0$ represents the IoU on clean inputs.

| Network | $\ell_\infty$ norm of perturbation, $\epsilon$ | | | | | | | | |
|---|---|---|---|---|---|---|---|---|---|
| | 0 | 0.25 | 0.5 | 1 | 2 | 4 | 8 | 16 | 32 |
| SegNet (VGG) | 43.0 | 32.3 | 25.9 | 19.5 | 14.8 | 11.7 | 9.7 | 6.9 | 4.0 |
| Deeplab v2 ASPP (VGG) | 66.9 | 55.3 | 44.1 | 31.7 | 22.5 | 17.2 | 13.9 | 11.8 | 9.1 |
| Dilated Frontend (VGG) | 67.1 | 56.7 | 45.7 | 33.8 | 24.2 | 19.2 | 16.1 | 12.2 | 8.2 |
| FCN8s (VGG) | 68.7 | 55.7 | 45.4 | 36.1 | 28.8 | 23.9 | 19.9 | 16.1 | 10.3 |
| FCN8s (ResNet) | 68.8 | 55.9 | 49.9 | 44.2 | 39.5 | 35.9 | 32.0 | 24.8 | 12.8 |
| Dilated Context (VGG) | 70.4 | 55.8 | 44.9 | 34.4 | 26.0 | 20.6 | 17.2 | 13.9 | 9.0 |
| Deeplab v2 ASPP (ResNet) | 73.3 | 61.6 | 52.7 | 43.3 | 35.9 | 30.7 | 27.7 | 24.6 | 18.5 |
| Deeplab v2 ASPP MS (ResNet) | 73.9 | **66.9** | 60.9 | **54.1** | **47.9** | **43.2** | **39.2** | **35.7** | **28.5** |
| PSPNet (ResNet) | **75.9** | 66.8 | 59.0 | 48.9 | 39.8 | 33.8 | 29.2 | 26.7 | 21.2 |

Table 8: The absolute IoU on the *Pascal VOC* dataset for various models when attacked with *FGSM ll*. This is evaluated for eight different values of the $\ell_\infty$ norm of the perturbation, $\epsilon$. $\epsilon = 0$ represents the IoU on clean inputs.

| Network | $\ell_\infty$ norm of perturbation, $\epsilon$ | | | | | | | | |
|---|---|---|---|---|---|---|---|---|---|
| | 0 | 0.25 | 0.5 | 1 | 2 | 4 | 8 | 16 | 32 |
| SegNet (VGG) | 43.0 | 36.2 | 27.4 | 17.6 | 11.4 | 8.3 | 6.7 | 5.4 | 4.1 |
| Deeplab v2 ASPP (VGG) | 66.9 | 61.5 | 52.3 | 34.6 | 17.3 | 9.5 | 7.0 | 6.1 | 5.6 |
| Dilated Frontend (VGG) | 67.1 | 61.6 | 51.9 | 35.8 | 19.1 | 10.6 | 6.6 | 5.5 | 4.4 |
| FCN8s (VGG) | 68.7 | 61.5 | 52.5 | 38.6 | 24.4 | 15.5 | 11.4 | 8.8 | 6.2 |
| FCN8s (ResNet) | 68.8 | 58.7 | 52.9 | 47.7 | 43.6 | **41.0** | **36.8** | 28.6 | 13.6 |
| Dilated Context (VGG) | 70.4 | 61.7 | 50.5 | 32.5 | 16.5 | 9.4 | 6.6 | 5.6 | 4.3 |
| Deeplab v2 ASPP (ResNet) | 73.3 | 67.8 | 60.4 | 49.1 | 37.5 | 30.0 | 25.7 | 22.0 | 17.2 |
| Deeplab v2 ASPP MS (ResNet) | 73.9 | **71.5** | **67.4** | **59.5** | **48.4** | 38.0 | 31.1 | 25.8 | 20.4 |
| PSPNet (ResNet) | **75.9** | 69.8 | 62.1 | 51.8 | 41.8 | 36.2 | 32.1 | **29.8** | **26.6** |

Table 9: The absolute IoU on the *Pascal VOC* dataset for various models when attacked with *Iterative FGSM*. This is evaluated for eight different values of the $\ell_\infty$ norm of the perturbation, $\epsilon$. $\epsilon = 0$ represents the IoU on clean inputs.

| Network | $\ell_\infty$ norm of perturbation, $\epsilon$ | | | | | | | | |
|---|---|---|---|---|---|---|---|---|---|
| | 0 | 0.25 | 0.5 | 1 | 2 | 4 | 8 | 16 | 32 |
| SegNet (VGG) | 43.0 | 32.3 | 25.9 | 21.3 | 12.7 | 8.9 | 5.6 | 3.8 | 2.8 |
| Deeplab v2 ASPP (VGG) | 66.9 | 55.3 | 44.1 | 40.3 | 26.0 | 12.5 | 7.6 | 4.7 | 3.4 |
| Dilated Frontend (VGG) | 67.1 | 56.7 | 45.7 | 42.1 | 24.4 | 13.4 | 9.1 | 5.6 | 4.1 |
| FCN8s (VGG) | 68.7 | 55.7 | 45.4 | 39.9 | 25.8 | 16.5 | 10.0 | 7.1 | 4.9 |
| FCN8s (ResNet) | 68.8 | 55.9 | 49.9 | 42.0 | 31.0 | 23.3 | 14.2 | 9.1 | 5.7 |
| Dilated Context (VGG) | 70.4 | 55.8 | 44.9 | 41.2 | 27.8 | 16.7 | 11.9 | 8.6 | 8.2 |
| Deeplab v2 ASPP (ResNet) | 73.3 | 61.6 | 52.7 | 47.3 | 32.2 | 22.1 | 13.1 | 7.5 | 5.3 |
| Deeplab v2 ASPP MS (ResNet) | 73.9 | **66.9** | **60.9** | **55.8** | **40.9** | **29.6** | **20.9** | **16.3** | **14.0** |
| PSPNet (ResNet) | **75.9** | 66.8 | 59.0 | 51.1 | 30.8 | 16.5 | 7.8 | 5.2 | 2.8 |

Table 10: The absolute IoU on the *Pascal VOC* dataset for various models when attacked with *Iterative FGSM ll*. This is evaluated for eight different values of the $\ell_\infty$ norm of the perturbation, $\epsilon$. $\epsilon = 0$ represents the IoU on clean inputs.

| Network | $\ell_\infty$ norm of perturbation, $\epsilon$ | | | | | | | | |
|---|---|---|---|---|---|---|---|---|---|
| | 0 | 0.25 | 0.5 | 1 | 2 | 4 | 8 | 16 | 32 |
| SegNet (VGG) | 43.0 | 36.2 | 27.4 | 22.0 | 11.4 | 6.7 | 5.3 | 4.1 | 3.7 |
| Deeplab v2 ASPP (VGG) | 66.9 | 61.5 | 52.3 | 49.0 | 28.0 | 12.1 | 6.7 | 5.8 | 4.8 |
| Dilated Frontend (VGG) | 67.1 | 61.6 | 51.9 | 49.1 | 27.8 | 10.8 | 5.4 | 4.0 | 3.7 |
| FCN8s (VGG) | 68.7 | 61.5 | 52.5 | 52.5 | 33.0 | 17.1 | 10.4 | 8.4 | 6.8 |
| FCN8s (ResNet) | 68.8 | 58.7 | 52.9 | 47.8 | 37.6 | 28.9 | **18.2** | **12.2** | **7.9** |
| Dilated Context (VGG) | 70.4 | 61.7 | 50.5 | 48.9 | 22.9 | 9.2 | 5.6 | 5.0 | 4.1 |
| Deeplab v2 ASPP (ResNet) | 73.3 | 67.8 | 60.4 | 56.9 | 39.6 | 21.1 | 11.3 | 7.7 | 6.3 |
| Deeplab v2 ASPP MS (ResNet) | 73.9 | **71.5** | **67.4** | **65.2** | **52.6** | **30.2** | 15.5 | 9.1 | 7.1 |
| PSPNet (ResNet) | **75.9** | 69.8 | 62.1 | 58.5 | 37.2 | 20.0 | 11.1 | 7.9 | 5.1 |

Table 11: The absolute IoU on the *Cityscapes* dataset for various models when attacked with *FGSM*. This is evaluated for eight different values of the $\ell_\infty$ norm of the perturbation, $\epsilon$. $\epsilon = 0$ represents the IoU on clean inputs.

| Network | \multicolumn{9}{c}{$\ell_\infty$ norm of perturbation, $\epsilon$} | | | | | | | | |
|---|---|---|---|---|---|---|---|---|---|
| | 0 | 0.25 | 0.5 | 1 | 2 | 4 | 8 | 16 | 32 |
| ENet | 53.4 | 39.6 | 35.6 | 31.0 | 24.0 | 13.2 | 5.8 | 4.1 | 1.4 |
| ICNet | 56.5 | 47.0 | 41.3 | 35.5 | 28.5 | 16.8 | 4.5 | 2.4 | 0.8 |
| FCN8 (VGG) | 62.1 | 46.0 | 38.0 | 31.9 | 27.8 | 23.9 | **16.2** | **7.7** | **3.9** |
| Dilated Frontend (VGG) | 59.0 | 46.3 | 38.1 | 31.1 | 25.7 | 20.7 | 13.3 | 5.0 | 1.7 |
| Dilated Context (VGG) | 62.3 | 48.4 | 39.0 | 31.6 | 26.0 | 20.8 | 13.3 | 4.8 | 1.8 |
| PSPNet (ResNet) | **74.4** | **58.5** | **52.9** | **48.9** | **46.0** | **36.3** | 16.0 | 2.8 | 1.9 |

Table 12: The absolute IoU on the *Cityscapes* dataset for various models when attacked with *FGSM ll*. This is evaluated for eight different values of the $\ell_\infty$ norm of the perturbation, $\epsilon$. $\epsilon = 0$ represents the IoU on clean inputs.

| Network | \multicolumn{9}{c}{$\ell_\infty$ norm of perturbation, $\epsilon$} | | | | | | | | |
|---|---|---|---|---|---|---|---|---|---|
| | 0 | 0.25 | 0.5 | 1 | 2 | 4 | 8 | 16 | 32 |
| ENet | 53.4 | 38.5 | 31.7 | 24.2 | 17.0 | 8.9 | 4.3 | 3.8 | 1.4 |
| ICNet | 56.5 | 47.2 | 40.5 | 33.2 | 25.1 | 13.4 | 3.4 | 2.3 | 0.8 |
| FCN8 (VGG) | 62.1 | 53.8 | 46.0 | 38.4 | 32.5 | 26.3 | 14.9 | **6.4** | **3.8** |
| Dilated Frontend (VGG) | 59.0 | 50.9 | 42.0 | 32.8 | 24.6 | 16.8 | 8.7 | 3.1 | 1.7 |
| Dilated Context (VGG) | 62.3 | 53.2 | 42.5 | 31.8 | 22.8 | 15.1 | 8.2 | 3.0 | 1.7 |
| PSPNet (ResNet) | **74.4** | **64.9** | **59.1** | **55.0** | **51.3** | **39.5** | **16.5** | 2.8 | 1.9 |

Table 13: The absolute IoU on the *Cityscapes* dataset for various models when attacked with *Iterative FGSM*. This is evaluated for eight different values of the $\ell_\infty$ norm of the perturbation, $\epsilon$. $\epsilon = 0$ represents the IoU on clean inputs.

| Network | \multicolumn{9}{c}{$\ell_\infty$ norm of perturbation, $\epsilon$} | | | | | | | | |
|---|---|---|---|---|---|---|---|---|---|
| | 0 | 0.25 | 0.5 | 1 | 2 | 4 | 8 | 16 | 32 |
| ENet | 53.4 | 39.6 | 35.6 | 22.6 | 14.2 | 9.3 | 5.7 | 3.6 | 2.7 |
| ICNet | 56.5 | 47.0 | 41.3 | 30.9 | 22.4 | 13.6 | 7.6 | 4.8 | 3.4 |
| FCN8 (VGG) | 62.1 | 46.0 | 38.0 | 31.1 | 19.1 | 11.1 | 5.8 | 4.0 | 3.2 |
| Dilated Frontend (VGG) | 59.0 | 46.3 | 38.1 | 30.6 | 18.1 | 10.0 | 5.4 | 3.5 | 2.8 |
| Dilated Context (VGG) | 62.3 | 48.4 | 39.0 | 31.4 | 18.1 | 9.6 | 5.1 | 3.4 | 2.7 |
| PSPNet (ResNet) | **74.4** | **58.5** | **52.9** | **40.2** | **25.4** | **16.4** | **8.9** | **5.7** | **4.3** |

Table 14: The absolute IoU on the *Cityscapes* dataset for various models when attacked with *Iterative FGSM ll*. This is evaluated for eight different values of the $\ell_\infty$ norm of the perturbation, $\epsilon$. $\epsilon = 0$ represents the IoU on clean inputs.

| Network | \multicolumn{9}{c}{$\ell_\infty$ norm of perturbation, $\epsilon$} | | | | | | | | |
|---|---|---|---|---|---|---|---|---|---|
| | 0 | 0.25 | 0.5 | 1 | 2 | 4 | 8 | 16 | 32 |
| ENet | 53.4 | 38.5 | 31.7 | 19.2 | 14.6 | 9.2 | 5.1 | 3.5 | 2.7 |
| ICNet | 56.5 | 47.2 | 40.5 | 33.8 | 22.4 | 14.5 | 8.9 | 6.8 | 5.5 |
| FCN8 (VGG) | 62.1 | 53.8 | 46.0 | 36.5 | 24.8 | 14.0 | 7.7 | 5.9 | 4.9 |
| Dilated Frontend (VGG) | 59.0 | 50.9 | 42.0 | 31.8 | 20.0 | 10.5 | 5.3 | 4.7 | 4.0 |
| Dilated Context (VGG) | 62.3 | 53.2 | 42.5 | 32.2 | 19.9 | 8.8 | 4.8 | 3.6 | 2.8 |
| PSPNet (ResNet) | **74.4** | **64.9** | **59.1** | **46.1** | **36.5** | **26.1** | **16.9** | **11.5** | **8.8** |

# D. Multiscale Processing and Transferability of Adversarial Examples

This section details additional results with both Deeplab v2 and FCN8s.

## D.1. Deeplab v2

Table 15 shows the performance, measured in IoU, on the VOC validation set when the input image is processed at different resolutions (50%, 75%, 100%). The fact that a different IoU is obtained for each input resolution, even though the weights of the network are the same, confirms that the network is not scale invariant. Note that the version of Deeplab which processes images at all the aforementioned resolutions, and max-pools the prediction at each pixel obtains the highest IoU. An alternative to max-pooling the predictions from each scale is to average-pool them. This method gives an insignificant improvement in accuracy, but does improve robustness as shown in Fig. 16.

Table 15: Performance of Deeplab v2 (ResNet) on the VOC validation set when processing images at different resolutions

| Model Name | IoU [%] |
| --- | --- |
| Deeplab v2 50% scale | 67.8 |
| Deeplab v2 75% scale | 71.9 |
| Deeplab v2 100% scale | 73.3 |
| Deeplab v2 100% scale (average pooling) | 73.4 |
| Deeplab v2 Multiscale (max pooling) | 73.9 |

### D.1.1 Average-pooling instead of max-pooling

As shown in Fig. 16, average-pooling the results from each scale is also more robust to all the adversarial attacks we tested compared to the single-scale version of Deeplab v2. In fact, multiscale processing (either max- or average-pooling) achieves a higher IoU Ratio at almost all $\epsilon$ values for each attack.

Table 17 also shows that black-box attacks generated from multiscale-averaging also transfer better to single scales of Deeplab v2, for all four adversarial attacks considered in this paper. This is similar to the case of max-pooling as shown in the main paper.

### D.1.2 Transferability experiments using the FGSM ll and Iterative FGSM attacks

Table 18 shows the transferability of adversarial attacks to different scales of Deeplab v2 using the FGSM ll and Iterative FGSM attacks. The main paper presented results using the FGSM and Iterative FGSM ll attacks. However, our findings remain consistent on these different attacks. The multiscale version of Deeplab v2 is the most robust to these attacks (as also seen in Fig. 14 and 16), and black-box attacks from it transfer the best to other scales of Deeplab v2.

Table 16: Performance of FCN8s when processing images at different resolutions. As with Deeplab v2, max-pooling the predictions from multiple scales achieves the best results.

| Model Name | IoU [%] |
| --- | --- |
| FCN8s 50% scale | 60.8 |
| FCN8s 75% scale | 67.8 |
| FCN8s 100% scale | 68.7 |
| FCN8s Multiscale | 69.9 |

### D.1.3 Transferability experiments at multiple $\epsilon$ values

Figure 17 shows the results of black-box attacks for multiple $\epsilon$ values between different scales of Deeplab v2 for the FGSM attack. The results are largely consistent with those at $\epsilon = 8$ as reported in the main paper – the multiscale version of Deeplab v2 is the most robust to white-box attacks and black-box attacks generated from it transfer the best to other scales of Deeplab v2. Also note how the transferability from each scale to another varies greatly. For example, attacks generated from the 50% scale transfer very poorly to 100% and vice versa.

## D.2. FCN8s

Table 16 shows the IoU of FCN8s (VGG) as the input resolution of the image is varied from the VOC dataset. As with Deeplab v2, a multiscale version which max-pools the predictions from each scale achieves the highest IoU.

The transferability experiments from Section 6 of the paper are repeated on FCN8 in Tables 19 and 20. Note that FCN8s has not been trained in a multiscale manner as Deeplab v2, and it is rather done as a post-processing step. Nevertheless, the results show a similar trend as Deeplab v2: The multiscale network is more robust to white-box attacks and black-box attacks generated from it transfer better. This suggests that training the network in a multiscale manner does not confer robustness to adversarial examples. Rather it is the fact that CNNs are not scale invariant, and that adversarial examples generated at one scale are not as malignant at another. Finally Fig. 18 shows the transferability experiments at multiple $\epsilon$ values, as was done for Deeplab v2 in the previous subsection.

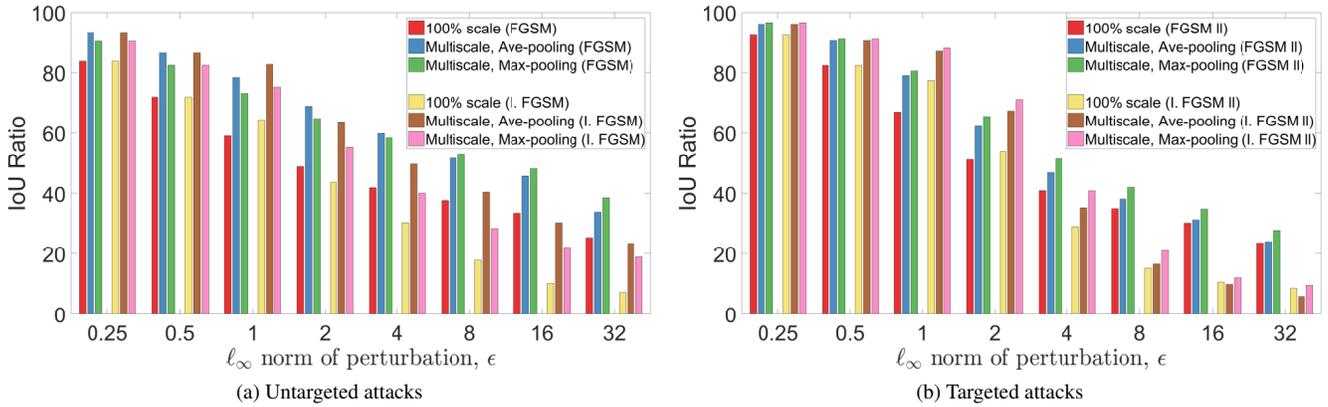

Figure 16: Adversarial robustness of Deeplab ASPP (single-scale) and Deeplab Multiscale ASPP. We compare two types of multiscale ensembling – max-pooling and average-pooling the predictions from each of the three scales of Deeplab v2 (ResNet 101). Note that both average- and max-pooling are more robust than just a single-scale model, achieving higher IoU Ratios for almost every $\epsilon$ value for each attack on the Pascal VOC dataset.

Table 17: Transferability of adversarial examples generated from different scales of Deeplab v2 (columns) and evaluated on different networks (rows). In this case, the outputs from each scale are *average-pooled* instead of max-pooled. The underlined diagonals for each attack show white-box attacks. Off-diagonals, show transfer (black-box) attacks. The most effective one in bold, is typically from the multiscale version of Deeplab v2. In the case of Iterative FGSM ll, black-box attacks from the multiscale networks are sometimes even more effective than white-box ones.

| Network evaluated | FGSM ($\epsilon = 8$) | | | | Iterative FGSM ll ($\epsilon = 8$) | | | |
|---|---|---|---|---|---|---|---|---|
| | 50% | 75% | 100% | Multiscale | 50% | 75% | 100% | Multiscale |
| Deeplab v2 0.5 (ResNet) | <u>37.3</u> | 70.5 | 84.8 | **48.8** | <u>18.0</u> | 92.0 | 96.9 | **12.1** |
| Deeplab v2 0.75 (ResNet) | 85.5 | <u>39.7</u> | 62.2 | **54.2** | 99.5 | <u>17.9</u> | 89.9 | **17.4** |
| Deeplab v2 1 (ResNet) | 93.6 | 57.9 | <u>37.7</u> | **51.7** | 100.0 | 79.0 | <u>15.5</u> | **9.6** |
| Deeplab v2 Multiscale (ResNet) | 75.1 | **54.2** | 59.0 | <u>51.6</u> | 95.2 | **84.9** | 87.5 | <u>16.7</u> |

| Network evaluated | FGSM ll ($\epsilon = 8$) | | | | Iterative FGSM ($\epsilon = 8$) | | | |
|---|---|---|---|---|---|---|---|---|
| | 50% | 75% | 100% | Multiscale | 50% | 75% | 100% | Multiscale |
| Deeplab v2 50% (ResNet) | <u>36.4</u> | 70.1 | 83.7 | **36.6** | <u>21.3</u> | 90.9 | 97.0 | **37.3** |
| Deeplab v2 75% (ResNet) | 89.9 | <u>37.4</u> | 61.6 | **39.9** | 99.1 | <u>20.0</u> | 88.6 | **44.1** |
| Deeplab v2 100% (ResNet) | 95.1 | 58.3 | <u>35.1</u> | **36.9** | 100.2 | 71.9 | <u>18.6</u> | **33.5** |
| Deeplab v2 Multiscale (ResNet) | 96.0 | **91.4** | 94.7 | <u>38.2</u> | 94.5 | **76.2** | 86.5 | <u>37.7</u> |

Table 18: Transferability of adversarial examples generated from different scales of Deeplab v2 (columns) and evaluated on different networks (rows). As with the main paper, *max-pooling* is performed from the output of each scale. However, in contrast to the main paper, the FGSM ll and Iterative FGSM attacks are reported. The underlined diagonals for each attack show white-box attacks. Off-diagonals, show transfer (black-box) attacks. The most effective one in bold, is typically from the multiscale version of Deeplab v2.

| Network evaluated | FGSM ll ($\epsilon = 8$) | | | | Iterative FGSM ($\epsilon = 8$) | | | |
|---|---|---|---|---|---|---|---|---|
| | 50% | 75% | 100% | Multiscale | 50% | 75% | 100% | Multiscale |
| Deeplab v2 0.5 (ResNet) | <u>36.4</u> | 70.1 | 83.7 | **46.0** | <u>21.3</u> | 90.9 | 97.0 | **39.2** |
| Deeplab v2 0.75 (ResNet) | 89.9 | <u>37.4</u> | 61.6 | **43.3** | 99.1 | <u>20.0</u> | 88.6 | **34.0** |
| Deeplab v2 1 (ResNet) | 95.1 | 58.3 | <u>35.1</u> | **33.9** | 100.2 | 71.9 | <u>18.6</u> | **22.0** |
| Deeplab v2 Multiscale (ResNet) | 90.7 | **60.8** | 68.9 | <u>42.1</u> | 96.5 | **81.9** | 87.5 | <u>29.2</u> |
| Deeplab v2 (VGG) | 95.1 | 69.9 | 63.8 | **61.9** | 98.5 | 86.9 | 86.3 | **81.2** |
| FCN8 (VGG) | 94.5 | 67.7 | 64.7 | **62.4** | 98.7 | 86.9 | 86.0 | **82.0** |

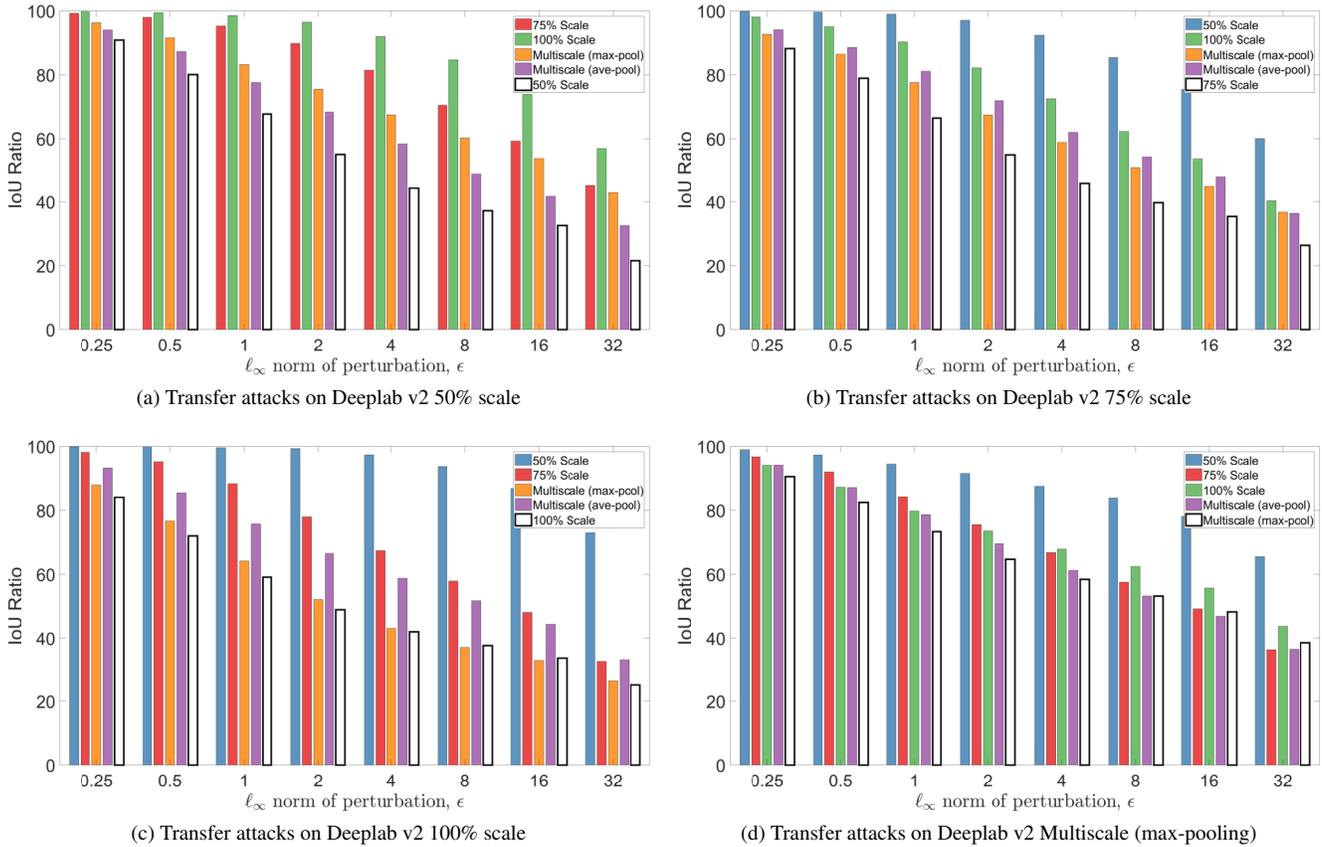

Figure 17: Black-box attacks on each scale of Deeplab v2, from each other scale, using adversarial perturbations generated by FGSM for differing values of $\epsilon$ on the Pascal VOC dataset. In each figure, the last bar shows the "white-box" attack on the network, where the attack is generated from the network that is being evaluated. This is typically the most powerful attack, as expected. Note that attacks generated from the multiscale version of Deeplab v2 (using either max- or average-pooling) produce the most effective black-box attacks across multiple $\epsilon$ values. The trend from the main paper, which only tabulated the IoU Ratio for $\epsilon = 8$, can thus be seen across all other $\epsilon$ values considered in this paper.

Table 19: Transferability of adversarial examples generated from different scales of FCN8s (VGG) (columns) and evaluated on different networks (rows) on the Pascal VOC dataset. For the multiscale network, the outputs from each scale are max-pooled. The underlined diagonals for each attack show white-box attacks. Off-diagonals, show transfer (black-box) attacks. The most effective one in bold, is typically from the multiscale version of FCN8s.

| Network evaluated | FGSM ($\epsilon = 8$) | | | | Iterative FGSM ll ($\epsilon = 8$) | | | |
|---|---|---|---|---|---|---|---|---|
| | 50% | 75% | 100% | Multiscale | 50% | 75% | 100% | Multiscale |
| FCN8 50% | <u>32.1</u> | **53.3** | 81.0 | 53.7 | <u>20.5</u> | 87.3 | 96.9 | **21.9** |
| FCN8 75% | 78.4 | <u>30.9</u> | 45.5 | **40.5** | 96.3 | <u>17.6</u> | 77.8 | **20.5** |
| FCN8 100% | 94.0 | 41.7 | <u>28.9</u> | **28.7** | 98.2 | 58.6 | <u>15.3</u> | **17.5** |
| FCN8 Multiscale | 79.1 | **42.8** | 53.3 | <u>47.8</u> | 97.5 | **79.3** | 85.2 | <u>20.0</u> |

Table 20: Transferability of adversarial examples generated from different scales of FCN8s (VGG) (columns) and evaluated on different networks (rows) on the Pascal VOC dataset. For the multiscale network, the outputs from each scale are max-pooled. The underlined diagonals for each attack show white-box attacks. Off-diagonals, show transfer (black-box) attacks. The most effective one in bold, is typically from the multiscale version of FCN8s.

| Network evaluated | FGSM ll ($\epsilon = 8$) | | | | Iterative FGSM ($\epsilon = 8$) | | | |
| --- | --- | --- | --- | --- | --- | --- | --- | --- |
| | 50% | 75% | 100% | Multiscale | 50% | 75% | 100% | Multiscale |
| FCN8 50% | <u>18.5</u> | 51.4 | 79.2 | **24.0** | <u>23.6</u> | 85.7 | 97.1 | **38.1** |
| FCN8 75% | 80.9 | <u>18.5</u> | 37.0 | **23.4** | 97.3 | <u>15.9</u> | 74.7 | **28.1** |
| FCN8 100% | 93.0 | 33.8 | <u>16.6</u> | **17.1** | 99.1 | 54.9 | <u>14.7</u> | **18.1** |
| FCN8 Multiscale | 87.5 | **40.0** | 60.3 | <u>21.1</u> | 96.4 | **74.5** | 82.3 | <u>25.1</u> |

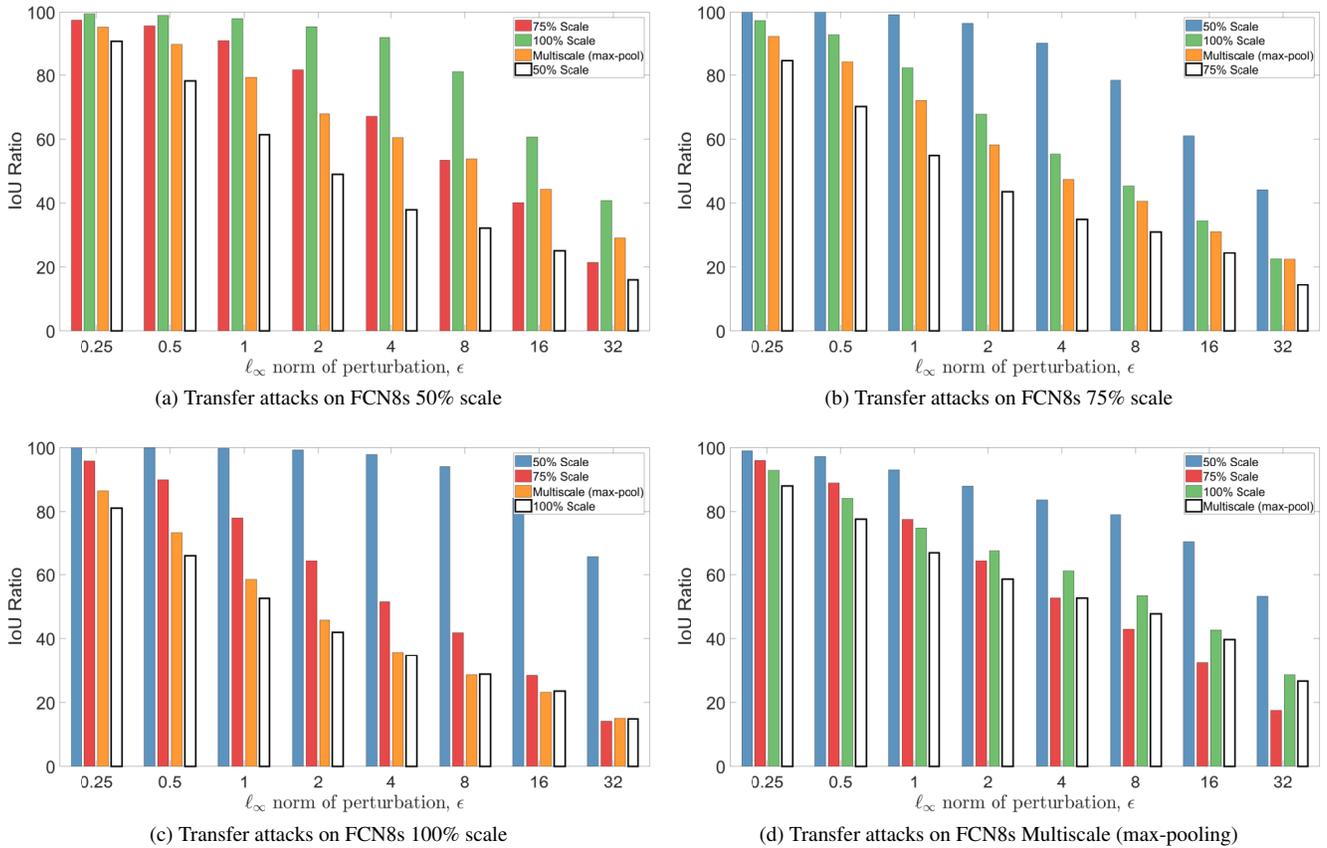

(a) Transfer attacks on FCN8s 50% scale

(b) Transfer attacks on FCN8s 75% scale

(c) Transfer attacks on FCN8s 100% scale

(d) Transfer attacks on FCN8s Multiscale (max-pooling)

Figure 18: Black-box attacks on each scale of FCN8, from each other scale, using adversarial perturbations generated by FGSM for differing values of $\epsilon$ on the Pascal VOC dataset. In each figure, the last bar shows the "white-box" attack on the network, where the attack is generated from the network that is being evaluated. The results from this experiment are very similar to Deeplab v2 – attacks generated from the multiscale network transfer the best to other scales. However, unlike Deeplab v2, the FCN8s network in this case was not trained with multiscale ensembling. This was simply done at test-time. This suggests that the increased robustness of multiscale networks to adversarial attacks, and their transferability to other networks, is not a result of the training procedure, but rather the fact that these networks are not scale invariant.

# E. Effect of CRFs on Adversarial Robustness

## E.1. Adversarial Robustness and Smoothing

The pairwise term of DenseCRF [52] (which is interpreted as a neural network in CRF-RNN [93]) takes the form of a weighted sum of a Bilateral and Gaussian filter.

$$\psi_p(x_i, x_j) = \mu(x_i, x_j) \left[ w_1 \exp\left(\frac{|p_i - p_j|^2}{\theta_\alpha} + \frac{|I_i - I_j|^2}{\theta_\beta}\right) + w_2 \exp\left(\frac{|p_i - p_j|^2}{\theta_\gamma}\right) \right]. \quad (8)$$

Increasing $\theta_\alpha$, $\theta_\beta$, $\theta_\gamma$, $w_1$ and $w_2$ all correspond to favouring smoother predictions. The compatibility function, $\mu(x_i, x_j)$, is given by the Potts model, and is equal to 1 if $x_i \neq x_j$ and 0 otherwise [52].

Figure 19 shows the effect of varying $\theta_\alpha$, Fig. 20 the effect of varying $\theta_\beta$ and Fig. 21 the effect of varying both $\theta_\gamma$ and $w_2$. Note that in all cases, each of the other hyperparameters remains unchanged at the values from the public CRF-RNN model.

In all of these cases, we can see that increasing the smoothness does not correspond to increasing adversarial robustness to the FGSM attack. Rather, as detailed in the next subsection, there is a correlation between the confidence of the prediction and robustness to the FGSM attack.

## E.2. Results about the confidence on VOC

We empirically measured the confidence of the predictions of CRF-RNN. This was done by recording the probability (from the softmax activation function) of the predicted (highest-scoring) label, and also by calculating the entropy of the marginal distribution over labels at each pixel in the image. A lower entropy indicates a more certain or confident prediction. This was then averaged over the Pascal VOC validation set.

Figures 22 and 23 show the mean confidence and entropy respectively as a function of the IoU Ratio. This is done for the FGSM attack for all the $\epsilon$ values considered in the paper. There is a clear correlation between the IoU Ratio and the confidence of the prediction. Moreover, the results of CRF-RNN are always more confident than FCN8s. Note that multiple variants of CRF-RNN, using different $\theta_\alpha$, $\theta_\beta$ and $\theta_\gamma$ hyperparameter values were considered, as in Figures 19 through 21.

## E.3. Experiments on Deeplab v2 with CRF

In contrast to CRF-RNN [93], a common approach is to apply CRFs as a post-processing step, as done in Deeplab [22]. We perform adversarial attacks on this by appending the CRF-RNN layer of [93] onto the Deeplab v2 network. This allows us to compute the gradient of the loss with respect to the input image (required for all the attacks) by backpropagating through the CRF-RNN layer. The parameters of the CRF-RNN layer appended to Deeplab v2 were manually set to the parameters used by the original authors[10] (who obtained them via cross-validation). Note that appending the CRF-RNN layer to Deeplab v2 and using the same parameters as the authors produces output that is identical to the post-processing code used by the original authors. The difference is that this allows us to compute gradients as well.

---

[10] http://liangchiehchen.com/projects/DeepLabv2_resnet.html

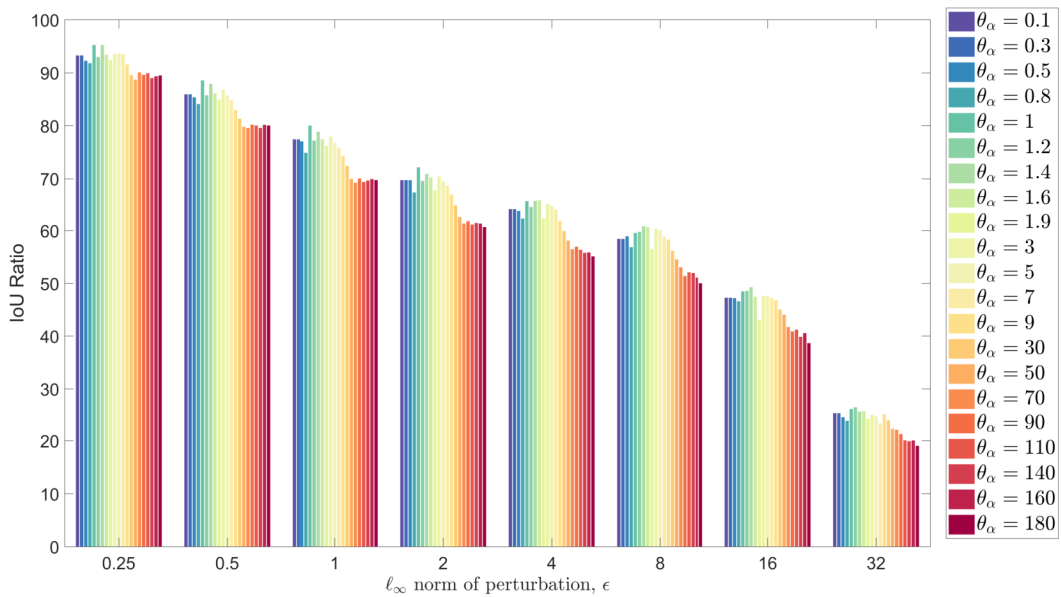

Figure 19: The IoU Ratio of CRF-RNN for various values of the $\theta_\alpha$ (filter bandwidth) hyperparameter when attacked with FGSM on the Pascal VOC dataset. Increasing this hyperparameter visually smoothes the result further, but we can see that this does not increase adversarial robustness. In fact, lower filter bandwidths of approximately $\theta_\alpha = 1$ provide more robustness.

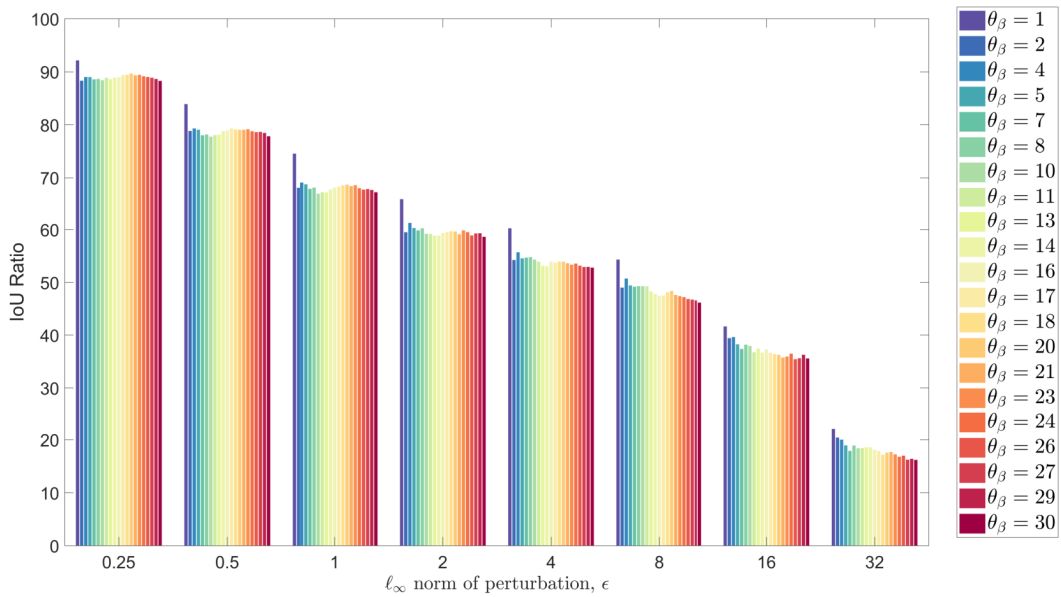

Figure 20: The IoU Ratio of CRF-RNN for various values of the $\theta_\beta$ (filter bandwidth) hyperparameter when attacked with FGSM on the Pascal VOC dataset. Again, we can see that larger filter bandwidths, which encourage more spatial smoothness, do not increase adversarial robustness.

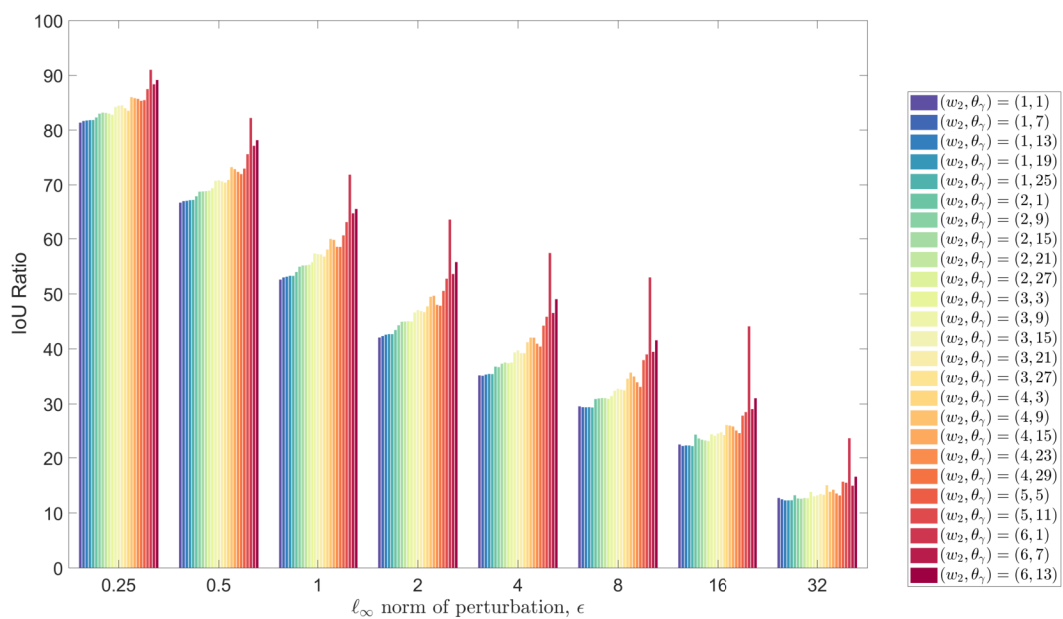

Figure 21: The IoU Ratio of CRF-RNN for various values of the $w_2$ and $\theta_\gamma$ parameters when attacked with FGSM on the Pascal VOC dataset. Increasing the weight of the Gaussian term ($w_2$) tends to increase robustness. However, we still see that lower filter bandwidths ($\theta_\gamma$) tend to provide more robustness.

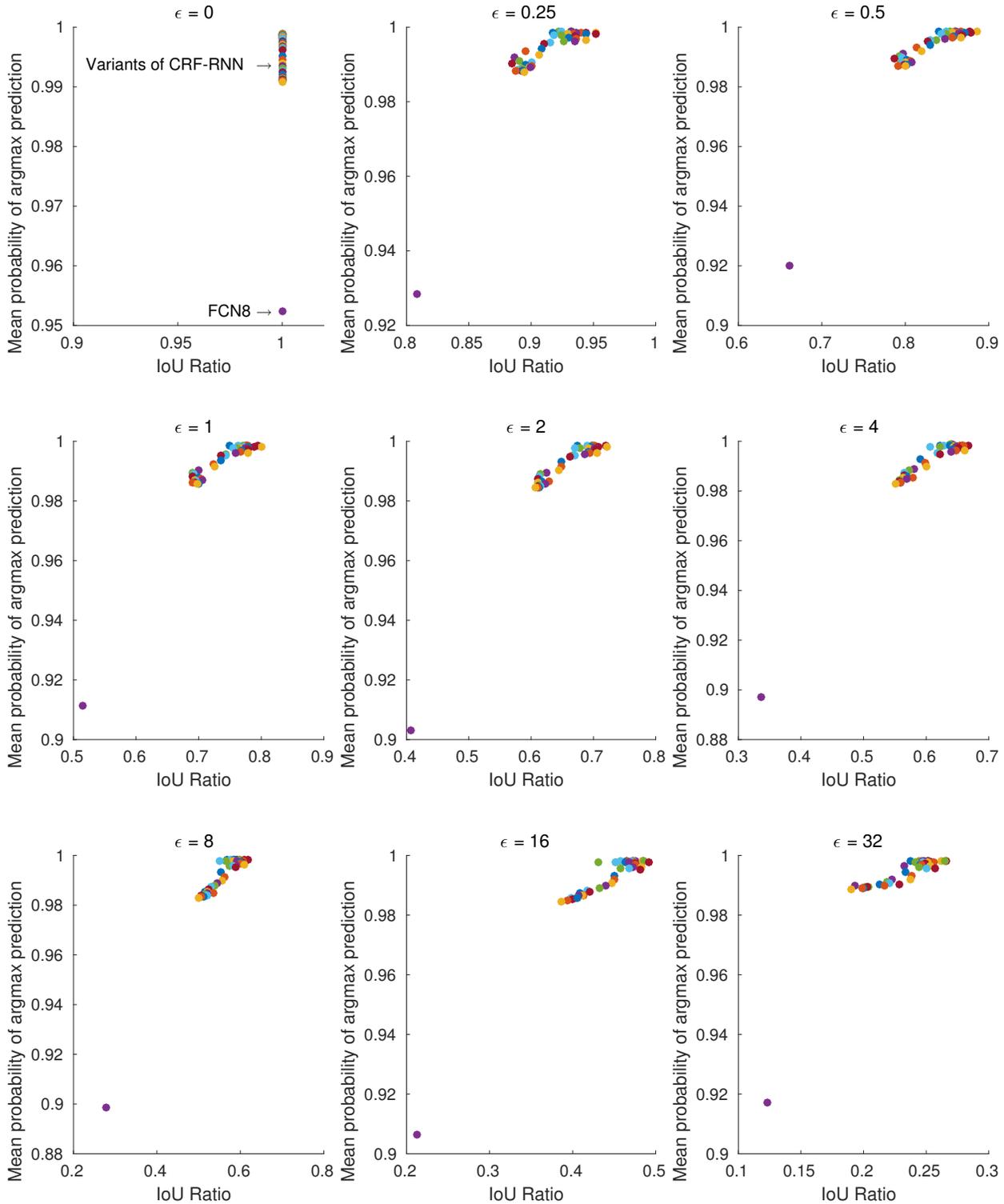

Figure 22: The mean probability of the highest-scoring class for each pixel, averaged over the Pascal VOC validation set. This is performed for the FGSM attack for multiple $\epsilon$ values. $\epsilon = 0$ corresponds to clean inputs (no adversarial attack). Note how FCN8s (the purple dot) consistently has the lowest mean probability. This probability is significantly lower than other variants of CRF-RNN (with varying $\theta_\alpha$, $\theta_\beta$, $\theta_\gamma$), shown by the other coloured dots. Moreover, note the correlation between the confidence in the prediction, and adversarial robustness to the FGSM attack. Additionally, the probability of the predicted class remains high (above 90%) for all models throughout all adversarial attacks.

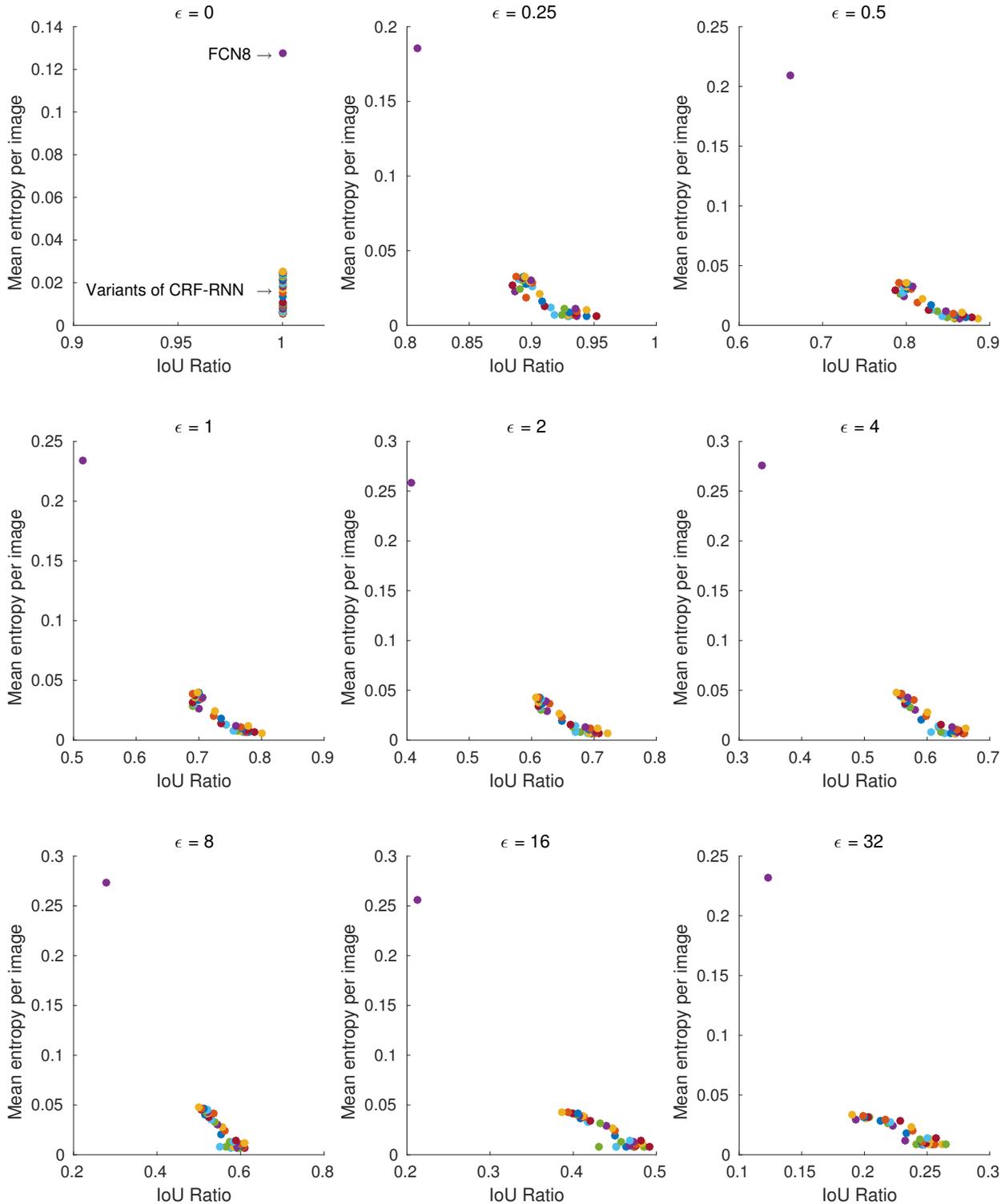

Figure 23: The mean entropy of the marginal distribution over all labels at each pixel, averaged over all images in the Pascal VOC validation set. A lower entropy corresponds to a more confident prediction. This is performed for the FGSM attack for multiple $\epsilon$ values. $\epsilon = 0$ corresponds to clean inputs (no adversarial attack). Note how FCN8s (the purple dot) consistently has the highest mean entropy (least confidence). This entropy is significantly higher than other variants of CRF-RNN (with varying $\theta_\alpha, \theta_\beta, \theta_\gamma$), shown by the other coloured dots. Moreover, note the correlation between the confidence in the prediction, and adversarial robustness to the FGSM attack.